\documentclass[twoside,11pt]{article}

\newcommand{\stydir}{.}
\newcommand{\bibsdir}{.}

\usepackage{amsmath,amssymb,amsfonts,graphicx,xcolor,nameref} 
\usepackage{subcaption}
\usepackage{\stydir/jmlr2e}

\usepackage{\stydir/jgmstd, \stydir/filtering, REFH}
\allowdisplaybreaks[1]

\renewcommand{\subcapref}[1]{(\protect\subref{subfig:#1})}

\newcommand{\FigTikzHMMs}{%
	\begin{figure}[!t]
		\begin{subfigure}[t]{0.5\textwidth}
			\centering
			\providecommand{\HMMinternode}{0.45in}
			\fbox{\includegraphics[width=0.95\textwidth]{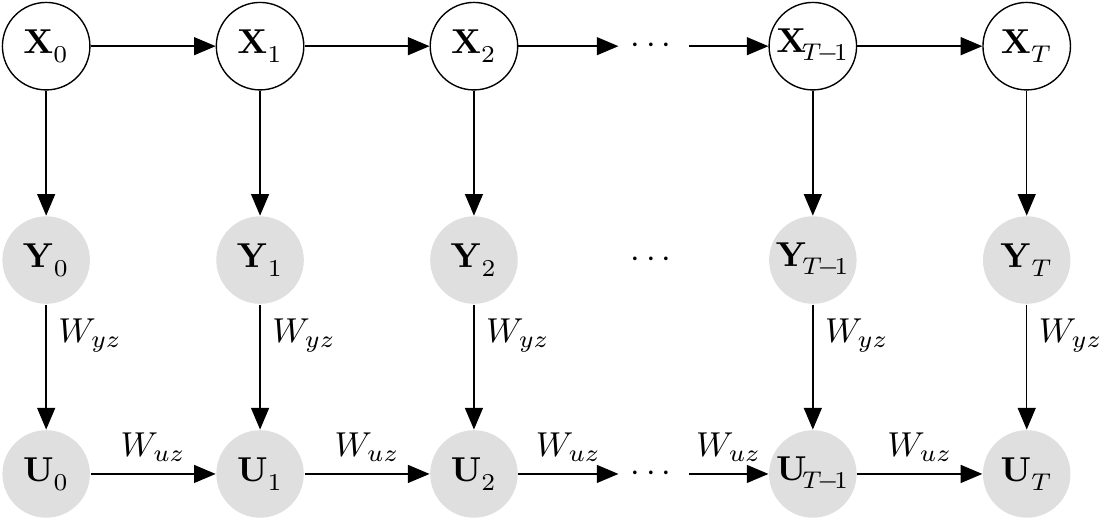}}
			\caption{}\label{subfig:HMMaugmented}
		\end{subfigure}
		\begin{subfigure}[t]{0.5\textwidth}
			\centering
			\providecommand{\HMMinternode}{0.45in}
			\fbox{\includegraphics[width=0.95\textwidth]{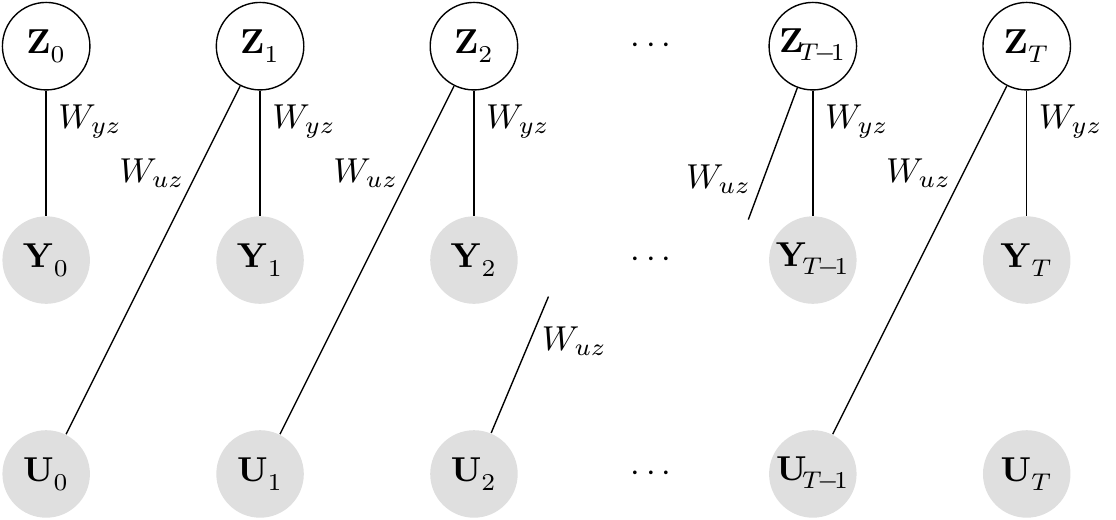}}
			\caption{}\label{subfig:REFHgraphicalmodel}
		\end{subfigure}
		\caption{
Two generative models for the observation sequences.
\subcapref{HMMaugmented} The data distribution. 
The bottom row is computed deterministically, 
$\protect\Rcrnts{t} := \xpct{q}{\protect\Modelstates{t}|\protect\stateobsvs{t},\protect\rcrnts{t-1};\params}$.
\subcapref{REFHgraphicalmodel} The rEFH.
At each time step $t$, an EFH with hidden layer $\protect\Modelstates{t}$ is trained on ``observations'' $\protect\Rcrnts{t-1}$ and $\protect\Stateobsvs{t}$.
The units in each layer are conditionally independent given the other layer, but the layers have been collapsed here to single nodes to save space (cf.\ \protect\subfig{EFHinrEFH}).
Weight matrices are labeled; biases are not shown.
	}\label{fig:tikzHMMs}
	\end{figure}
}

\newcommand{\FigTikzEFHs}{%
	\begin{figure}[!t]
		\providecommand{\NNinternode}{0.009in}%
		\providecommand{\NNinterlayer}{0.6in}%
		\providecommand{\NNnodesize}{0.2in}%
		\begin{subfigure}[t]{0.5\textwidth}
			\centering%
			\includegraphics[width=0.9\textwidth]{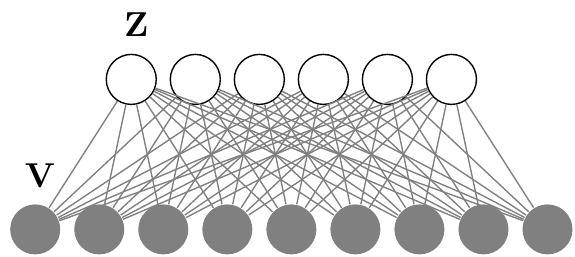}%
			\caption{}\label{subfig:EFH}%
		\end{subfigure}
 		\begin{subfigure}[t]{0.5\textwidth}
			\centering%
			\includegraphics[width=0.9\textwidth]{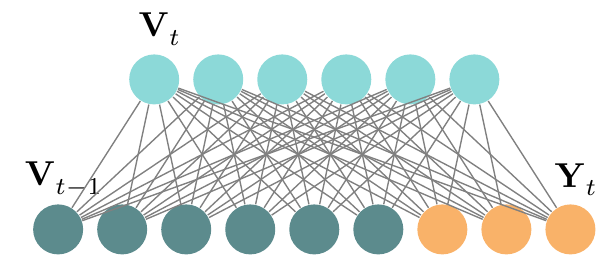}%
			\caption{}\label{subfig:EFHinrEFH}%
		\end{subfigure}
		\caption{
\subcapref{EFH} A graphical model for a generic EFH.
\subcapref{EFHinrEFH} The EFH composing time step $t$ of the rEFH, i.e., one time slice in \protect\subfig{REFHgraphicalmodel}.
Each input vector consists of the sufficient statistics for the previous hidden-layer ($\protect\rcrnts{t-1} = \protect\modelpostmeans{t-1}$, \RCRNTcolor) and the current observations ($\protect\stateobsvs{t}$, \PROPcolor).
}\label{fig:tikzEFH}
	\end{figure}
}

\newcommand{\FigTemporalEFHs}{
	\begin{figure} [!t]
		\captionsetup[subfigure]{aboveskip=0.73in}
		\begin{subfigure}[t]{0.5\textwidth}
			\vskip 0pt
			\centering%
			\providecommand{\numHid}{4}%
			\fbox{\includegraphics[width=0.95\textwidth]{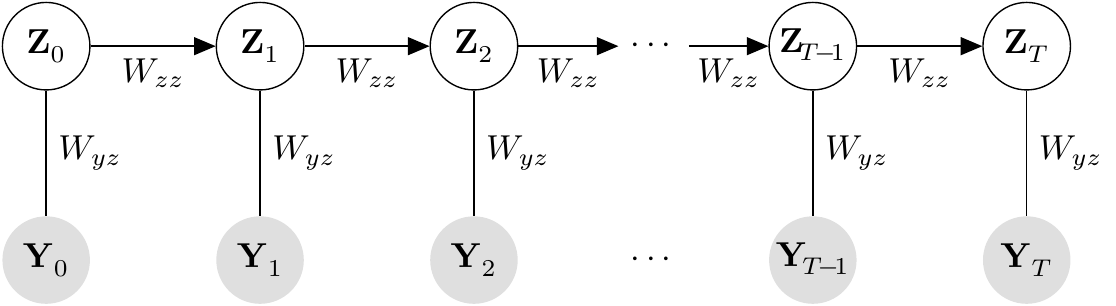}}%
			\caption{}\label{subfig:TRBMgraphicalmodel}%
		\end{subfigure}
		\captionsetup[subfigure]{aboveskip=0.15in,belowskip=0in}
		\begin{subfigure}[t]{0.5\textwidth}
			\vskip 0pt
			\fbox{\includegraphics[width=0.95\textwidth]{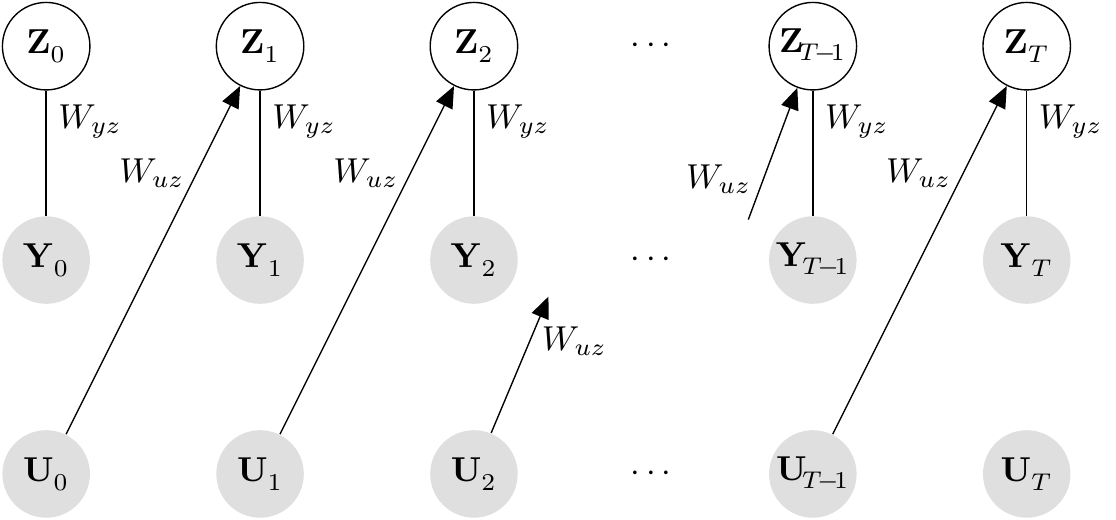}}%
			\caption{}\label{subfig:RTRBMgraphicalmodel}%
		\end{subfigure}
		\caption{Temporal extensions of the EFH.
Each node represents a layer, within which there are no connections, but which has been collapsed to save space.
\subcapref{TRBMgraphicalmodel} The non-approximated TRBM, to which learning rules are never directedly applied.
\subcapref{RTRBMgraphicalmodel} The TRBM and RTRBM.
At each time step $t$, an EFH with dynamical bias $\protect\rcrnts{t-1} = \protect\modelpostmeans{t-1}$ is trained on observations $\protect\stateobsvs{t}$.
}\label{fig:SutskeverModels}
	\end{figure}
}

\newcommand{\FigErrorStatsVsNumHiddens}{
	\begin{figure} [!t]
		\includegraphics[width=0.95\textwidth]{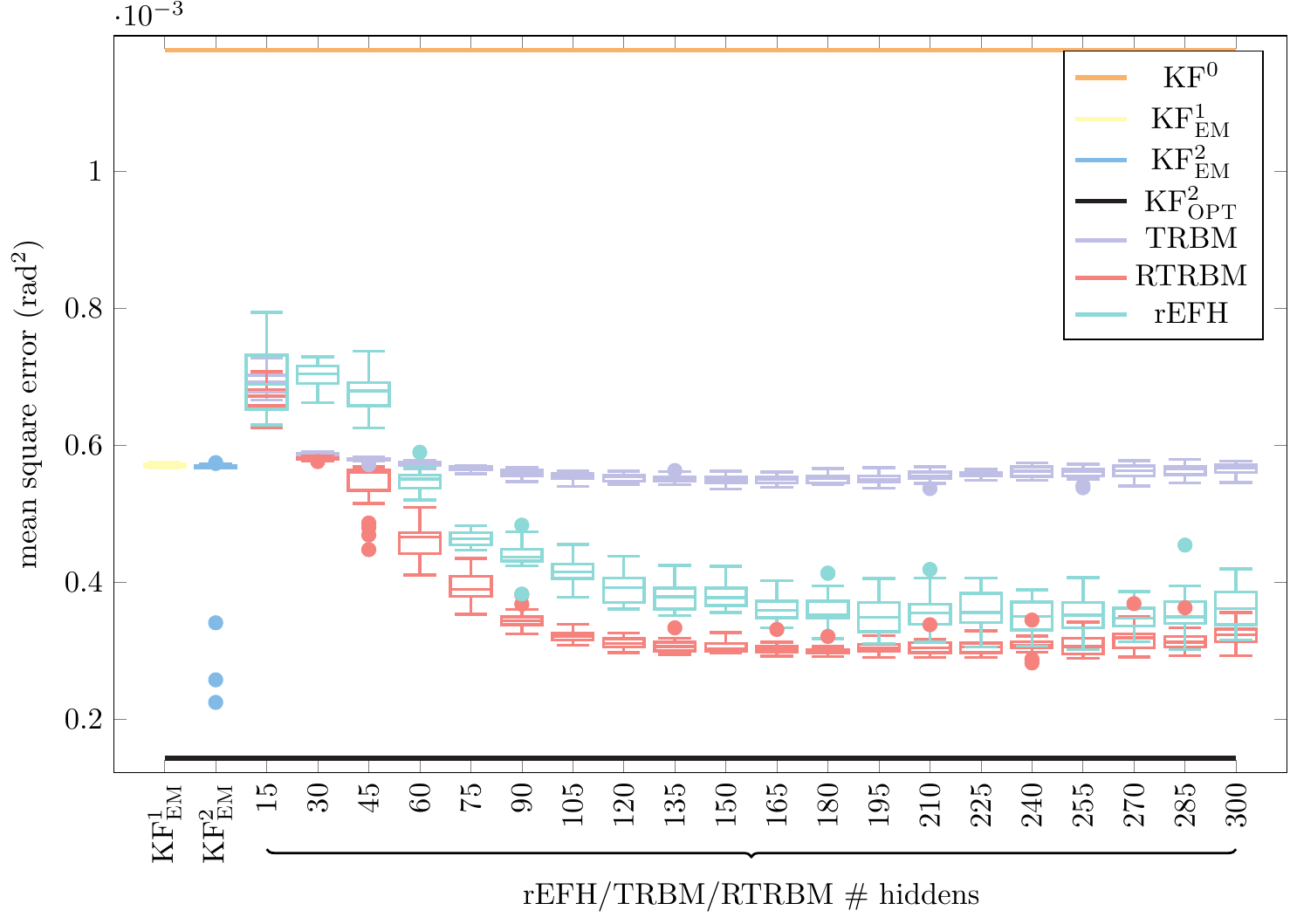}%
		\caption{Box-and-whisker plots of mean square errors (MSEs) for the rEFH, TRBM, RTRBM, and benchmark filters, across 20 repeated training instances, on a single testing set.
Median MSE for each set of networks is marked with a line; the box contains the interquartile range; whiskers extend to $1.5\times$ the interquartile range, beyond which outliers are marked with filled circles.
Performance of the benchmark filters \KFnaught and \KFopt\ is marked with long horizontal lines:\ no boxplot is possible, because these models require no training.
We have reported previously similar results for the rEFH and the benchmarks \citep{Makin2015b}.
}\label{fig:ErrorStatsVsNumHiddens}
	\end{figure}
}

\newcommand{\FigTikzWorldModel}{
	\begin{figure}[!t]
		\begin{subfigure}[t]{0.5\textwidth}
			\centering
			\includegraphics[width=0.53\textwidth]{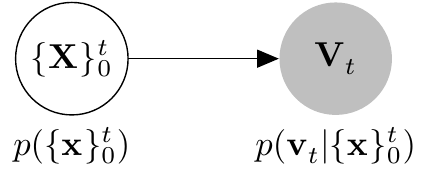}%
			\caption{}\label{subfig:simpleWorld}
		\end{subfigure}
		\begin{subfigure}[t]{0.5\textwidth}
			\centering
			\includegraphics[width=0.5\textwidth]{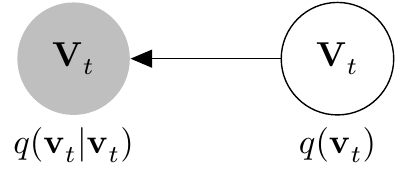}%
			\caption{}\label{subfig:simpleModel}
		\end{subfigure}
		\caption{Probabilistic graphical models.  
Observed nodes are shaded.
Training the model ($q$) makes the marginals match:\ $p(\protect\augobsvs{t}) = q(\protect\augobsvs{t})$.
\subcapref{simpleWorld} The ``world's'' generative process.
\subcapref{simpleModel} The model's generative process.
		}\label{fig:tikzWorldModel}
	\end{figure}
}

\newcommand{\FigTikzWorldModelWithProxcauses}{
	\begin{figure}[!t]
		\begin{subfigure}[t]{0.5\textwidth}
			\centering
			\includegraphics[width=0.5\textwidth]{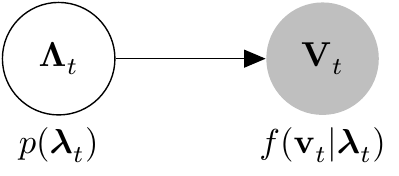}%
			\caption{}\label{subfig:simpleWorldWithProxcauses}
		\end{subfigure}
		\begin{subfigure}[t]{0.5\textwidth}
			\centering
			\includegraphics[width=0.5\textwidth]{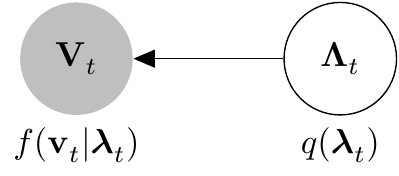}%
			\caption{}\label{subfig:simpleModelWithProxcauses}
		\end{subfigure}
		\caption{The models from \protect\fig{tikzWorldModel} are rewritten in terms of the common conditional distribution $f(\protect\augobsvs{t}|\protect\proxcauses{t})$ and the prior distributions $p(\protect\proxcauses{t})$ and $q(\protect\proxcauses{t})$ that are induced by the changes of variables $\protect\Proxcauses{t} := \protect\worldstateSSfunc{\protect\Worldstatestillnow}$ and $\protect\Proxcauses{t} := \modelstateSSfunc{\protect\Modelstates{t}}$, respectively.
\subcapref{simpleWorld} The ``world's'' generative process.
\subcapref{simpleModel} The model's generative process.
		}\label{fig:tikzWorldModelWithProxcauses}
	\end{figure}
}

\jmlrheading{1}{2015}{1-48}{9/15}{[[pub date]]}{Joseph G.\ Makin, Benjamin K.\ Dichter, and Philip N.\ Sabes}

\ShortHeadings{The Recurrent Exponential Family Harmonium}{Makin, Dichter, and Sabes}
\firstpageno{1}

\begin{document}

\title{Recurrent Exponential-Family Harmoniums without Backprop-through-Time}

\author{\name Joseph G.\ Makin \email makin@phy.ucsf.edu \\
		\name Benjamin K.\ Dichter\email ben.dichter@berkeley.edu \\
		\name Philip N.\ Sabes\email sabes@phy.ucsf.edu\\
       \addr Center for Integrative Neuroscience\\
       University of California, San Francisco\\
       San Francisco, CA 94143-0444, USA
       }

\editor{[[editor's name]]}

\maketitle

\begin{abstract}
Exponential-family harmoniums (EFHs), which extend restricted Boltzmann machines (RBMs) from Bernoulli random variables to other exponential families \citep{Welling2004}, are generative models that can be trained with unsupervised-learning techniques, like contrastive divergence \citep{Hinton2006a,Hinton2002}, as density estimators for static data.
Methods for extending RBMs---and likewise EFHs---to data with temporal dependencies have been proposed previously \citep{Sutskever2007,Sutskever2009}, the learning procedure being validated by qualitative assessment of the generative model.
Here we propose and justify, from a very different perspective, an alternative training procedure, proving sufficient conditions for optimal inference under that procedure.
The resulting algorithm can be learned with only forward passes through the data---backprop-through-time is not required, as in previous approaches.
The proof exploits a recent result about information retention in density estimators \citep{Makin2015a}, and applies it to a ``recurrent EFH'' (rEFH) by induction.
Finally, we demonstrate optimality by simulation, testing the rEFH:\ (1) as a filter on training data generated with a linear dynamical system, the position of which is noisily reported by a population of ``neurons'' with Poisson-distributed spike counts; and (2) with the qualitative experiments proposed by \citet{Sutskever2009}.
\end{abstract}


\section{Introduction}
We are interested in unsupervised learning of data with temporal dependencies.
The aim of such learning is to extract the statistical structure from ``trajectories'' or sequences,
after which it will be possible, for example, to predict the next frame in the sequence, to make optimal inferences about the dynamical state of the underlying system being observed (``filter''), or to generate novel sequences with the same statistical properties.

Ideally, we would make use of some of the sophisticated unsupervised-learning techniques developed for static data.
If, however, a single training vector consists of the observations at a single time step, the temporal dependencies obviously will not be learned.
The opposite extreme, feasible in some contexts, is to assign to a single training vector an entire trajectory of observations; but in general, trajectories may be very long, vary in length, or required to be learned on-line.

A middle course is to try to construct, at each time step, \emph{augmented} observations that are shorter than an entire trajectory of observations, but that form, unlike the raw observations, a Markov chain.
When the \emph{direct} temporal dependencies are not, indeed, the length of the trajectory itself, past observations' bearing on future observations can be summarized compactly.
Data consisting of the current observation, augmented by such a summary, form a Markov chain across time, which facilitates the application of recursive procedures for filtering and prediction.
In the most familiar example of this strategy, the Kalman filter, the past noisy observations of a discrete-time, linear-Gaussian dynamical system are summarized by the posterior cumulants over a hidden dynamical state.
These cumulants, which are also the sufficient statistics for the hidden state (conceived as a parameter), are the state-estimate vector, whose length is proportional to the length of the direct temporal dependencies, and a covariance matrix that encodes the (inverse) reliability of this state estimate.
The recursive filtering procedure in turn makes parameter learning possible---in this case, with an expectation-maximization (EM) algorithm \citep[see, e.g.,][]{Ghahramani1996a}.

Our goal is to extend this ``middle'' strategy to more general temporal sequences, for which we are unable, for example, to compute analytically the posterior distribution over the hidden state, or even to formulate equations of state or an observation model.
The compact summaries, consequently, we shall require the algorithm itself to formulate, rather than deriving them analytically.
In particular, our basic building block will be the exponential-family harmonium (EFH) \citep{Welling2004}, a two-layer Markov random field with full interlayer connections and no intralayer connections, in which the conditional distributions over the hidden layer given the visible layer and vice versa are products over exponential-family distributions.
(The restricted Boltzmann machine corresponds to the special case of Bernoulli conditional distributions.)
For the summaries, with which we augment the observations at each time step, we choose the hidden-unit means from the previous time step.

We have recently \citep{Makin2015b} introduced this temporal extension to the EFH, the ``recurrent EFH'' (rEFH); investigated its predictions for receptive-field formation in primate cerebral cortex; and compared its performance as a filter for simple dynamical systems against Kalman filters learned with EM.
Here, we investigate the analytical properties of the rEFH, supplying sufficient conditions and a proof for when the training procedure results in a model that provides optimal inference.
The central idea behind the proof is to abandon the assumption of model correctness that underlies use of, e.g., linear-Gaussian dynamical systems, in favor of a weaker assumption about model identifiability.
This assumption, which holds for the rEFH, guarantees that, at each time step, the sufficient statistics for the hidden units of the trained model are also sufficient statistics for the current \emph{true} hidden state.
That is, the model's hidden units learn to encode the compact summary corresponding to the filter cumulants.

The rEFH is very similar to, but importantly different from, two recent temporal extensions to the EFH, the TRBM \citep{Sutskever2007} and the RTRBM \citep{Sutskever2009}.
We show why the proof does not apply to these models, and then compare all three empirically by training and testing on data where the optimal posterior estimate of the true dynamical state is computable analytically.
We also compare model performance on the synthetic data set used in those papers:\ the rEFH is superior to the TRBM and competitive with the RTRBM, despite a simpler learning algorithm.
Finally, we show that generating sample trajectories---\emph{in the reverse direction}---is also simple in the rEFH, in contrast to the TRBM and RTRBM, for which generation is hard.

\section{Theory}\label{sec:theory}
We begin by describing our strategy for extending the EFH to temporal data (\sctn{proofstrategy}) before justifying it more rigorously (\sctn{proofformal}), and finally describing the training procedure in detail (\sctn{rEFHtraining}).
Our aim is a training procedure and architecture that can be guaranteed to provide \emph{optimal inference}.
That is, after training, our model should allow us to compute the posterior distribution, $p(\worldstates{t}|\stateobsvstillnow)$, over the dynamical state, $\Worldstates{t}$, of the underlying causes of our observations, $\stateobsvs{t}$, conditioned on the preceding sequence of such observations.
In particular, we require that the (mean) hidden-unit activities of the model at time $t$ encode the cumulants of this distribution.

\subsection{Strategy for extending the EFH to temporal data}\label{sec:proofstrategy}
We consider temporal data $\Stateobsvsalltime$ that have been generated by a very general discrete-time dynamical process, anything consistent with the independence statements asserted by the hidden Markov model in the top two rows of \subfig{HMMaugmented}, without any other assumptions about the distributions parameterizing it.
(The Markov assumption is weak in the sense that longer temporal dependencies can be accommodated under it simply by lengthening the state vector.)
We shall, however, introduce below some constraints relating to a technical condition called ``identifiability.''
Our goal is to make optimal inferences about the underlying ``causes,'' $\Worldstatesalltime$, for the observations, $\Stateobsvsalltime$, with joint distribution
\begin{equation}\nonumber
	p(\worldstatesalltime,\stateobsvsalltime)
		= \prod_{t=0}^T p(\stateobsvs{t}|\worldstates{t}) p(\worldstates{t}|\worldstates{t-1}),
\end{equation}
corresponding to \subfig{HMMaugmented}.
(Here as elsewhere, for notational compactness, $p(\worldstates{0}|\worldstates{-1}) := p(\worldstates{0})$, and likewise for other expressions that reference negative times.)
Given nothing more than samples from a ``data distribution'' of this general form, we construct our own latent-variable model, $q$, and attempt to bring the probabilities it assigns to observation sequences close to the probabilities with which they occur in the data; i.e., we seek parameters $\params$ such that $q(\stateobsvsalltime;\params) = p(\stateobsvsalltime)$ (``density estimation'').

\FigTikzHMMs

Here we introduce an important distinction.
We do not assume that the latent variables of the model have the same form as those ``in the world,'' $\Worldstates{}$, so in fact we use a different letter for them, $\Modelstates{}$.
In training latent-variable density estimators, the correspondence between $\Modelstates{}$ and $\Worldstates{}$ is not always straightforward, even after the model has become a good one for the observed data.
Nevertheless, our recent result about the character of this correspondence \citep{Makin2015a} can be used to show how to achieve our goal, optimal inference of the true causes, $\Worldstates{}$.

Consider in particular observed (static) random variables $\Augobsvs{}$.
In many cases one can say that the sufficient statistics\footnote{``T,'' the usual symbol for sufficient statistics, is already used for time, so we use ``U'' instead.} for the model's latent variables, $\SSformodelstates{}$, are also sufficient statistics for the true latent variables, $\Worldstates{}$.
This is intuitive:\ the same aspects of $\Augobsvs{}$ are in need of explanation, whether via the world or via the model.
The result does not hold for all latent-variable density estimators, but for a class of such models that obeys a set of rather weak technical conditions \citep[see][and \thm{maintheorem} below]{Makin2015a}, it is the case that
\begin{equation}\label{eqn:sufficiencyForLatents}
	\forall \augobsvs{}\hspace{0.1in} q(\augobsvs{};\params) = p(\augobsvs{}) 
	\implies \MI{\SSformodelstates{}}{\Worldstates{}} = \MI{\Augobsvs{}}{\Worldstates{}},
\end{equation}
with $\MI{\cdot}{\cdot}$ the mutual information.
That is, after a good model has been learned (left-hand equation), transforming $\Augobsvs{}$ into any set of sufficient statistics for $\Modelstates{}$ discards no information about the underlying cause, $\Worldstates{}$.
When information about $\Worldstates{}$ reaches $\Modelstates{}$ \emph{only} via $\Augobsvs{}$, \eqn{sufficiencyForLatents} is equivalent \citep[see][]{Makin2015a} to
\begin{equation}\label{eqn:posteriorMatchingResult}
	\forall \augobsvs{}\hspace{0.1in} q(\augobsvs{};\params) 
		= p(\augobsvs{}) 
	\implies \forall \augobsvs{},\worldstates{}\hspace{0.1in}
			p(\worldstates{}|\ssformodelstates{})
		= p(\worldstates{}|\augobsvs{}).
\end{equation}
In words, transforming $\Augobsvs{}$ into sufficient statistics for $\Modelstates{}$ does not change the posterior distribution over the underlying cause, $\Worldstates{}$.
For EFHs, the sufficient statistics $\SSformodelstates{}$ are the posterior mean, $\xpct{q}{\Modelstates{}|\Augobsvs{}}$; hence, in a trained model, the mean activity of the hidden units captures all the information about the true cause, $\Worldstates{}$, as was contained in the vector of observed responses, $\Augobsvs{}$.

We construct a procedure for training the EFH on data with temporal dependencies that exploits this result.
In particular, at every time step, we train the EFH on an ``augmented'' input vector $\augobsvs{t}$ that consists of the current observations, $\stateobsvs{t}$, concatenated together with the sufficient statistics for $\Modelstates{t-1}$ from the previous time step, $\ssformodelstates{t-1}$---i.e., the previous posterior mean, $\xpct{q}{\Modelstates{t-1}|\augobsvs{t-1}}$.
The central intuition for why this procedure will produce models providing optimal inference is as follows.
Training the model at the first time step ($t=0$) forces the hidden units---or, more precisely, the sufficient statistics for them, $\xpct{q}{\Modelstates{0}|\stateobsvs{0}}$---to extract from observations $\stateobsvs{0}$ a good summary of the underlying state, $\Worldstates{0}$, no information about it being lost.
The sufficient statistics at the second time step ($t=1$), then, are forced to become a good ``meta-summary'' of both the information in $\Stateobsvs{1}$ about $\Worldstates{1}$ \emph{and} the previous summary.
Continuing in this fashion builds up in these sufficient statistics---the means of the hidden units---a good running summary about the true state and its history.
We call the resulting model the recurrent, exponential-family harmonium (rEFH).

\subsection{Rigorous justification of the training procedure}\label{sec:proofformal}
We now formalize this intuitive argument.
We begin with the result from our previous work:
\begin{theorem}
Let $p(\augobsvs{}) = \sum_\worldstates{} p(\augobsvs{}|\worldstates{})p(\worldstates{})$ and 
$q(\augobsvs{};\params) = \sum_\modelstates{} p(\augobsvs{}|\modelstates{};\params)p(\modelstates{};\params)$
be expressible as \emph{identifiable} finite mixtures of the same conditional distribution $f(\augobsvs{}|\proxcauses{})$, 
$\sum_\proxcauses{} f(\augobsvs{}|\proxcauses{})p(\proxcauses{})$ and
$\sum_\proxcauses{} f(\augobsvs{}|\proxcauses{})q(\proxcauses{})$, respectively.
That is, the map from prior distributions over ``parameters'' $\Proxcauses{}$ to mixtures (marginal distributions over $\Augobsvs{}$) is injective.
Then \eqn{sufficiencyForLatents} holds.
Furthermore, if $\Worldstates{}$ and $\Modelstates{}$ are independent conditioned on $\Augobsvs{}$, then \eqn{posteriorMatchingResult} holds as well.
\end{theorem}
For the proof, see the work of \citet{Makin2015a}.
Here we prove by induction that the training procedure just proposed generalizes this theorem to temporal data.
\begin{theorem}\label{thm:maintheorem}
For all $t$, let the marginal distributions
\begin{equation}\nonumber
	\begin{split}
		p(\augobsvs{t};\params) 
			&:= \sum_{\worldstatestillnow;\params}p(\augobsvs{t}|\worldstatestillnow)p(\worldstatestillnow)\\
		q(\augobsvs{t};\params) 
			&:= \sum_\modelstates{t} q(\augobsvs{t}|\modelstates{t};\params)q(\modelstates{t};\params)
	\end{split}
\end{equation}
be expressible as \emph{identifiable} finite mixtures of the same conditional distribution $f(\augobsvs{t}|\proxcauses{t})$, 
$\sum_\proxcauses{t} f(\augobsvs{t}|\proxcauses{t})p(\proxcauses{t})$ and
$\sum_\proxcauses{t} f(\augobsvs{t}|\proxcauses{t})q(\proxcauses{t})$, respectively.
Furthermore, let the distribution of observed data $\Augobsvs{t}$ be consistent with the independence statements asserted by \subfig{HMMaugmented}, in which in particular each observation consists of the emissions of a hidden Markov model, concatenated together with the sufficient statistics, $\SSformodelstates{t-1}$, for the latent variables $\Modelstates{t-1}$ of $q$ at the preceding time step:
\begin{equation}\label{eqn:augmentedData}
	\Augobsvs{t} := [\SSformodelstates{t-1},\Stateobsvs{t}].
\end{equation}
Then
\begin{equation}\label{eqn:sufficiencyForTemporalLatents}
	\forall t, \forall \augobsvs{t}\hspace{0.1in} q(\augobsvs{t};\params) = p(\augobsvs{t};\params) 
	\implies \MI{\SSformodelstates{t}}{\Worldstates{t}} = \MI{\Stateobsvstillnow}{\Worldstates{t}}.
\end{equation}
If, moreover, $\Worldstates{t}$ and $\Modelstates{t}$ are independent conditioned on $\Augobsvs{t}$, then 
\begin{equation}\label{eqn:posteriorGivenHiddensIsFilteringDistribution}
	\forall t, \forall \augobsvs{t}\hspace{0.1in} q(\augobsvs{t};\params) = p(\augobsvs{t};\params) 
	\implies p(\worldstates{t}|\ssformodelstates{t}) = p(\worldstates{t}|\stateobsvstillnow). 
\end{equation}
\end{theorem}

Note that the variables $\ssformodelstates{t-1}$ can be calculated deterministically from $\stateobsvstillprev$, so the data distribution $p$ induces a distribution over these sufficient statistics as well as the observations.
This is the joint distribution we have called $p(\augobsvs{t};\params)$, whose generative model is depicted in \subfig{HMMaugmented}.
The equality on the left of \eqn{sufficiencyForTemporalLatents} says that the EFH is a good density estimator for its inputs at time $t$, $\Augobsvs{t}$ (cf.\ \subfig{REFHgraphicalmodel}).
The information equality on the right says that all the information about the current state, $\Worldstates{t}$, that was available in all the observations up to the present, $\stateobsvstillnow$, is also available in a particular function, $\suffstats{\Modelstates{t}}(\cdot)$, of just the current augmented input vector, $\augobsvs{t}$.
Just as \eqn{sufficiencyForLatents} can be re-expressed as \eqn{posteriorMatchingResult}, this information equality can be alternatively expressed as \eqn{posteriorGivenHiddensIsFilteringDistribution}.

The critical technical condition guaranteeing \eqn{sufficiencyForLatents}, on which the proof depends, is that $p(\augobsvs{t})$ and $q(\augobsvs{t})$ be expressible in terms of the same finite-mixture model, and that the map from distributions over parameters of this mixture to the marginal distribution over $\Augobsvs{t}$ be injective, except possibly on sets of measure zero.
Precise characterization being rather technical, we defer it to the \app{technicalconditions}.
We now prove theorem \thm{maintheorem} by induction, under the assumption that these technical conditions hold.

\begin{proof}
It is easily verified that $\Worldstates{t}$ and $\Modelstates{t}$ are independent conditioned on $\Augobsvs{t}$, for all time, so we shall work throughout with \eqns{posteriorMatchingResult}{posteriorGivenHiddensIsFilteringDistribution} rather than \eqns{sufficiencyForLatents}{sufficiencyForTemporalLatents}.
Now, the base case of the induction is simply an application of \eqn{posteriorMatchingResult}.
Assume that $q(\augobsvs{0};\params) = p(\augobsvs{0})$, or equivalently that $q(\stateobsvs{0};\params) = p(\stateobsvs{0})$ (since at $t=0$ there is no recurrent vector to concatenate with the current observation $\stateobsvs{0}$; see \subfig{REFHgraphicalmodel}); that is, that the EFH has become a good density estimator for the initial data.
Then \eqn{posteriorMatchingResult} implies that
\begin{equation}\label{eqn:filteringInductionBaseCase}
	p(\worldstates{0}|\ssformodelstates{0})
		= p(\worldstates{0}|\suffstats{\Modelstates{0}}(\stateobsvs{0})) 
		= p(\worldstates{0}|\stateobsvs{0}).
\end{equation}

Now for the induction step, assume that
\begin{equation}\label{eqn:filteringInductionIf}
	p(\worldstates{t-1}|\ssformodelstates{t-1}) = p(\worldstates{t-1}|\stateobsvstillprev).
\end{equation}
We want to show that under this assumption, the EFH having become a good density estimator for its inputs at time $t$, $q(\augobsvs{t};\params) = p(\augobsvs{t};\params)$, implies
\begin{equation}\label{eqn:filteringInductionThen}
	p(\worldstates{t}|\ssformodelstates{t}) = p(\worldstates{t}|\stateobsvstillnow).
\end{equation}
Together with the base case, this will show that the implication holds for all time---\eqn{posteriorGivenHiddensIsFilteringDistribution}.

To save space we abbreviate $\Rcrnts{t} := \SSformodelstates{t}$ and omit references to the parameters.
Now, since $q(\augobsvs{t}) = p(\augobsvs{t})$, \eqn{posteriorMatchingResult} implies that
\begin{subequations}
	\begin{align}
		p(\worldstates{t}|\rcrnts{t})
			&= 	p(\worldstates{t}|\augobsvs{t}) \nonumber\\
			&= 	p(\worldstates{t}|\stateobsvs{t},\rcrnts{t-1}) \nonumber\\
	\implies p(\worldstates{t}|\rcrnts{t}) p(\stateobsvs{t}|\rcrnts{t-1})
			&= 	p(\worldstates{t}|\stateobsvs{t},\rcrnts{t-1})
				p(\stateobsvs{t}|\rcrnts{t-1}) \label{eqn:rEFHproofSequenceA}\\
			&= 	p(\stateobsvs{t}|\worldstates{t},\rcrnts{t-1})
				p(\worldstates{t}|\rcrnts{t-1}) \nonumber\\
			&= 	p(\stateobsvs{t}|\worldstates{t}) \sum_{\worldstates{t-1}} 
				p(\worldstates{t},\worldstates{t-1}|\rcrnts{t-1}) \label{eqn:rEFHproofSequenceB}\\
			&= 	p(\stateobsvs{t}|\worldstates{t}) \sum_{\worldstates{t-1}} 
				p(\worldstates{t}|\worldstates{t-1},\rcrnts{t-1})
				p(\worldstates{t-1}|\rcrnts{t-1}) \nonumber\\
			&= 	p(\stateobsvs{t}|\worldstates{t}) \sum_{\worldstates{t-1}} 
				p(\worldstates{t}|\worldstates{t-1}) 
				p(\worldstates{t-1}|\rcrnts{t-1}) \label{eqn:rEFHproofSequenceC}\\
			&= 	p(\stateobsvs{t}|\worldstates{t}) \sum_{\worldstates{t-1}}
				p(\worldstates{t}|\worldstates{t-1},\stateobsvstillprev) 
				p(\worldstates{t-1}|\stateobsvstillprev) \label{eqn:rEFHproofSequenceD}\\
			&= 	p(\stateobsvs{t}|\worldstates{t})
				p(\worldstates{t}|\stateobsvstillprev) \nonumber\\
			&= 	p(\stateobsvs{t}|\worldstates{t},\stateobsvstillprev)	
				p(\worldstates{t}|\stateobsvstillprev) \label{eqn:rEFHproofSequenceE}\\
			&= 	p(\worldstates{t}|\stateobsvstillnow)
				p(\stateobsvs{t}|\stateobsvstillprev). \label{eqn:rEFHproofSequenceF}
	\end{align}
\end{subequations}
The independence conditions used in the transitions to 
\eqnsss{rEFHproofSequenceB}{rEFHproofSequenceC}{rEFHproofSequenceD}{rEFHproofSequenceE}
are licensed by the graph in \subfig{HMMaugmented}, in particular by the ``explaining away'' property of the unobserved $\Rcrnts{t}$.
Marginalizing out $\worldstates{t}$ from the left-hand side of \eqn{rEFHproofSequenceA} and from \eqn{rEFHproofSequenceF} shows that 
that $p(\stateobsvs{t}|\rcrnts{t-1}) = p(\stateobsvs{t}|\stateobsvstillprev)$,
and therefore $p(\worldstates{t}|\rcrnts{t}) = p(\worldstates{t}|\stateobsvstillnow)$, which is \eqn{filteringInductionThen}.
This completes the induction and therefore the proof of the theorem.
\end{proof}

\subsection{Training the rEFH}\label{sec:rEFHtraining}
The proof shows that optimal inference---what is expressed by the information equality on the right side of \eqn{sufficiencyForTemporalLatents}---can be achieved, in density estimators that can be expressed as identifiable mixture models, by establishing the equality on the left side, $q(\augobsvs{t};\params) = p(\augobsvs{t};\params)$,
where $\augobsvs{t}$ is constructed according to \eqn{augmentedData}.
To establish that equality, we train an exponential-family harmonium \citep[EFH, depicted in \subfig{EFH}, and introduced by][]{Welling2004} on these ``augmented'' data vectors.
The joint distribution then takes the form of a Boltzmann distribution.
For simplicity, we consider only exponential families whose sufficient statistics are the identity function.
In accordance with \eqn{augmentedData}, the visible layer consists of the current observations, $\Stateobsvs{t}$, and the previous hidden-layer sufficient statistics, $\Rcrnts{t-1}$; see \subfig{EFHinrEFH}.
Thus the joint distribution is
\begin{equation}\label{eqn:REFHjointdstrb}
	q(\modelstates{t},\rcrnts{t-1},\stateobsvs{t};\params) 
		\propto h(\rcrnts{t-1})h(\stateobsvs{t})h(\modelstates{t})\exp\big\{ 
			\modelstates{t}\tr (\vishidwts \stateobsvs{t} + \rcrnthidwts\rcrnts{t-1}  + \hidbiases)
			+ \visbiases\tr\stateobsvs{t} + \rcrntbiases\tr\rcrnts{t-1}\big\}.
\end{equation}
Here the functions $h$ are the base measures, determined by the choices of exponential families.
The intractable normalizer for $q$ has been omitted.
The corresponding conditional distributions are%
\begin{subequations}\label{eqn:REFHconditionals}%
	\begin{align}
		q(\modelstates{t}|\rcrnts{t-1},\stateobsvs{t};\params) 
			&= \frac{1}{\mathcal{Z}_\modelstatevar(\stateobsvs{t},\rcrnts{t-1})}h(\modelstates{t})
				\exp\big\{ (\vishidwts\stateobsvs{t} + \rcrnthidwts\rcrnts{t-1} + \hidbiases)\tr\modelstates{t} \big\},
				\label{eqn:REFHhiddencond}\\
		q(\stateobsvs{t}|\modelstates{t};\params) 
			&= \frac{1}{\mathcal{Z}_\stateobsvvar(\modelstates{t})}h(\stateobsvs{t})
				\exp\big\{ (\vishidwts\tr\modelstates{t} + \visbiases)\tr\stateobsvs{t} \big\},
				\label{eqn:REFHstateobsvcond}\\
		q(\rcrnts{t-1}|\modelstates{t};\params) 
			&= \frac{1}{\mathcal{Z}_\rcrntvar(\modelstates{t})}h(\rcrnts{t-1})
				\exp\big\{ (\rcrnthidwts\tr\modelstates{t} + \rcrntbiases)\tr\rcrnts{t-1} \big\}.
				\label{eqn:REFHrcrntcond}
	\end{align}
\end{subequations}
The conditional distribution for the visible units has been separated into two conditionals, \eqns{REFHstateobsvcond}{REFHrcrntcond}, corresponding to the observations and the previous hidden-layer means, respectively.
From the graphical model, \subfig{EFHinrEFH}, their product forms the conditional distribution over the visible units:\
$q(\rcrnts{t},\stateobsvs{t}|\modelstates{t};\params) = q(\rcrnts{t}|\modelstates{t};\params) q(\stateobsvs{t}|\modelstates{t};\params)$.

The data distribution over $\Stateobsvsalltime$ is assumed to be unknown, but the conditional distribution it induces over the hidden-layer sufficient statistics is known and deterministic:
\begin{equation}\label{eqn:rcrntdstrb}
	p(\rcrnts{t}|\rcrnts{t-1},\stateobsvs{t};\params)
		:= \delta(\rcrnts{t} - \xpct{q}{\Modelstates{t}|\rcrnts{t-1},\stateobsvs{t};\params}).
\end{equation}
In words, the distribution over the current hidden-layer sufficient statistics is a delta function located at the posterior mean.
In the EFH, this is computed as a deterministic ``upward'' (visible-to-hidden) pass through the weight matrix, followed by the pointwise nonlinearity appropriate for the exponential family to which the posterior distribution belongs \citep[see][]{Welling2004}.
We emphasize that, despite being parameterized (and by parameters of the model, no less), we treat the conditional distribution in \eqn{rcrntdstrb} as part of the data distribution ($p$) because it provides part of the training data for the EFH.

\FigTikzEFHs

Exact density estimation in an EFH---that is, gradient descent in the KL divergence between $p(\augobsvs{t};\params)$ and $q(\augobsvs{t};\params)$---requires prolonged Gibbs sampling, which introduces variance into the learning procedure as well as slowing it down \citep{Hinton2002}.
We therefore proceed by stochastic gradient descent of the approximate ``$n$-step contrastive divergence'' (CD$_n$) objective function, which has $q = p$ at its minimum \citep{Hinton2002,Hinton2006a}.
At the cost of this approximation, contrastive-divergence learning buys fast, low-variance learning, since it requires only $n$ (which can be small) full steps of Gibbs sampling from the model before the weights can be updated.
We also ignore the dependence of the data distribution, $p(\augobsvs{t};\params)$, on the parameters, introduced by \eqn{rcrntdstrb}, 
an approximation that we discuss further in \sctn{conclusions}, below.

The learning rules are therefore the standard rules for the EFH under one-step contrastive divergence \citep[see][]{Welling2004}, applied to our data:
\begin{align}\label{eqn:rEFHparameterUpdates}
	\Delta\vishidwts &\propto \sum_{t=0}^T 
		\smplavg{
			q(\modelstates{t}|\rcrnts{t-1},\stateobsvs{t})
			p(\rcrnts{t-1},\stateobsvs{t})}
			{\Modelstates{t}\Stateobsvs{t}\tr
			-\smplavg{
				q(\updtmodelstates{t}|\updtrcrnts{t-1},\updtstateobsvs{t})
				q(\updtrcrnts{t-1},\updtstateobsvs{t}|\modelstates{t})}
				{\Updtmodelstates{t}\Updtstateobsvs{t}\tr
			}
		},\nonumber \\
	\Delta\rcrnthidwts &\propto \sum_{t=0}^T 
		\smplavg{
			q(\modelstates{t}|\rcrnts{t-1},\stateobsvs{t})
			p(\rcrnts{t-1},\stateobsvs{t})}
			{\Modelstates{t}\Rcrnts{t-1}\tr
			-\smplavg{
				q(\updtmodelstates{t}|\updtrcrnts{t-1},\updtstateobsvs{t})
				q(\updtrcrnts{t-1},\updtstateobsvs{t}|\modelstates{t})}
				{\Updtmodelstates{t}\Updtrcrnts{t-1}\tr}
		},\nonumber \\
	\Delta\hidbiases &\propto \sum_{t=0}^T 
		\smplavg{
			q(\modelstates{t}|\rcrnts{t-1},\stateobsvs{t})
			p(\rcrnts{t-1},\stateobsvs{t})}
			{\Modelstates{t} 
			-\smplavg{
				q(\updtmodelstates{t}|\updtrcrnts{t-1},\updtstateobsvs{t})
				q(\updtrcrnts{t-1},\updtstateobsvs{t}|\modelstates{t})}
				{\Updtmodelstates{t}}
			},\nonumber \\
	\Delta\visbiases &\propto \sum_{t=0}^T 
		\smplavg{
			q(\modelstates{t}|\rcrnts{t-1},\stateobsvs{t})
			p(\rcrnts{t-1},\stateobsvs{t})}
			{\Stateobsvs{t}
			-\smplavg{
				q(\updtstateobsvs{t}|\modelstates{t})}
				{\Updtstateobsvs{t}}
			},\nonumber \\
	\Delta\rcrntbiases &\propto \sum_{t=0}^T 
		\smplavg{
			q(\modelstates{t}|\rcrnts{t-1},\stateobsvs{t})
			p(\rcrnts{t-1},\stateobsvs{t})}
			{\Rcrnts{t-1}
			-\smplavg{
				q(\updtrcrnts{t-1}|\modelstates{t})}
				{\Updtrcrnts{t-1}}
			},
\end{align}
where the angle brackets indicate sample averages rather than true expectations.

\section{Alternative extensions of the EFH to temporal data.}
Sutskever and colleagues introduced an extension of the restricted Boltzmann machine (RBM) to temporal data, and proposed two different methods for training it \citep{Sutskever2007,Sutskever2009}.
We re-present them here to contrast them with the rEFH, and to show that the proof of the preceding section does not apply to these methods.
For simplicity, we confine ourselves to the special case of the TRBM where the only direct dependence of an RBM on any of its predecessors in the chain is on the \emph{hidden} units of its \emph{immediate} predecessor (\subfig{TRBMgraphicalmodel}).
(We consider the more general model in the discussion below.)
We derive both models in terms of an EFH, rather than an RBM as in the original papers, because the EFH is more general and can be accommodated within the framework of the cited works with minor adjustments only.
Finally, our development of the models is slightly different, although the final result is the same.

\citet{Sutskever2007} define the following joint distribution:
\begin{equation}\nonumber
	q(\modelstates{0},\stateobsvs{0},\ldots,\modelstates{T},\stateobsvs{T};\params) 
		= \prod_{t=0}^T q(\modelstates{t},\stateobsvs{t}|\modelstates{t-1};\params),
\end{equation}
reflecting the independence properties of the graph in \subfig{TRBMgraphicalmodel}.
The graph also asserts the independence of $\Stateobsvs{t}$ and $\Modelstates{t-1}$, given $\Modelstates{t}$.
This follows, straightforwardly, from the definition of the central component of the model,
\begin{equation}\nonumber
	q(\modelstates{t},\stateobsvs{t}|\modelstates{t-1};\params) 
		:= \frac{1}{\mathcal{Z}(\modelstates{t-1})}h(\stateobsvs{t})h(\modelstates{t})\exp\big\{ 
			\modelstates{t}\tr \vishidwts \stateobsvs{t} 
			+ \visbiases\tr\stateobsvs{t} 
			+ (\rcrnthidwts\modelstates{t-1}  
			+ \hidbiases)\tr\modelstates{t} 
			\big\},
\end{equation}
since the components of the right-hand side that reference $\modelstates{t-1}$ and $\stateobsvs{t}$ can be factored into separate pieces.
This equation has the form of a harmonium; that is, it can be produced by two conditional distributions of the form
\begin{subequations}\nonumber
	\begin{align}
		q(\modelstates{t}|\modelstates{t-1},\stateobsvs{t}) 
			&= \frac{1}{\mathcal{Z}_\modelstatevar(\stateobsvs{t},\modelstates{t-1})}h(\modelstates{t})
				\exp\big\{ (\vishidwts\stateobsvs{t} + \rcrnthidwts\modelstates{t-1}  + \hidbiases)\tr\modelstates{t} \big\},\\
		q(\stateobsvs{t}|\modelstates{t},\modelstates{t-1}) 
			&= \frac{1}{\mathcal{Z}_\stateobsvvar(\modelstates{t})}h(\stateobsvs{t})
				\exp\big\{ (\vishidwts\tr\modelstates{t} + \visbiases)\tr\stateobsvs{t} \big\}
			= q(\stateobsvs{t}|\modelstates{t}).
	\end{align}
\end{subequations}
As it stands, learning in this model is intractable, since inference does not admit of a recursive procedure (like the Kalman filter).
In order apply such a procedure, the authors decouple the EFHs through time with a gross approximation.
They define a recursion $\rcrnts{t} = \xpct{q}{\Modelstates{t}|\rcrnts{t-1},\stateobsvs{t}}$, where the expectation is taken under the first conditional.
Then they simply substitute $\rcrnts{t-1}$ for $\modelstates{t-1}$.
Thus, the approximate model is defined by the joint distribution
\begin{equation}\label{eqn:RTRBMjointdstrb}
	q(\modelstates{t},\stateobsvs{t}|\rcrnts{t-1};\params) 
		:= \frac{1}{\mathcal{Z}(\rcrnts{t-1})}h(\stateobsvs{t})h(\modelstates{t})\exp\big\{ 
			\modelstates{t}\tr \vishidwts \stateobsvs{t} 
			+ \visbiases\tr\stateobsvs{t} 
			+ (\rcrnthidwts\rcrnts{t-1}  
			+ \hidbiases)\tr\modelstates{t}\big\},
\end{equation}
corresponding to the graphical model in \subfig{RTRBMgraphicalmodel};
and the augmented data distribution is, as for the rEFH, defined by \eqn{rcrntdstrb}.
This is the defining equation for the TRBM and the RTRBM.

Applying the substitution ($\rcrnts{t-1}$ for $\modelstates{t-1}$) to the conditional distributions yields precisely \eqn{REFHhiddencond} and \eqn{REFHstateobsvcond}.
This makes the deterministic recursion the same for both models, to wit, \eqn{rcrntdstrb}.
Thus, the only difference between the (R)TRBM and the rEFH is that the latter additionally defines a conditional distribution over the previous hidden means, \eqn{REFHrcrntcond}.
This can also be seen in the joint distributions:\ the EFH defined in \eqn{RTRBMjointdstrb} is \emph{conditioned} on $\rcrnts{t-1}$, rather than defining a (joint) distribution over it, as in \eqn{REFHjointdstrb}.
\emph{The (R)TRBM does not define a distribution ($q$) over} $\Rcrnts{t}$, although the recursion extends the \emph{data} distribution ($p$) to $\Rcrnts{t}$.
The consequence can be seen in the subtle but important difference in the graphical models:\ the connections from $\Rcrnts{t-1}$ to $\Modelstates{t}$ are directed in the (R)TRBM (\subfig{RTRBMgraphicalmodel}), rather than undirected as in the rEFH (\subfig{REFHgraphicalmodel}).
Essentially, the vector $\Rcrnts{t-1}$ is treated in the (R)TRBM like a (dynamic) bias rather than data, as in the rEFH; it is not allowed to ``fantasize'' about the past, only the present.
This leads to different learning rules, which we exhibit below.

Treating the past hidden-unit activities as data is, however, required for our proof that the rEFH provides optimal inference (\sctn{proofformal}).
Recall the basic idea:
The hidden-unit activities of a(n identifiable) generative model become good summaries for the data they are required to generate; if the past hidden-unit activities are among those data, the summaries accumulate recursively.
It is not obvious, therefore, that such a proof can be extended to the (R)TRBM (the left-hand equality in \eqn{sufficiencyForTemporalLatents} will not be satisfied).
Nevertheless, the RTRBM does appear to achieve nearly optimal inference, as we show in our experiments below.
But choosing to treat the previous hidden-unit activities as data (the rEFH) does turn out to have a very important consequence:\ the dependence of these ``data'' on the rEFH's parameters, introduced by \eqn{rcrntdstrb}, can be safely ignored.
When the recurrent activities are treated as a dynamic bias, on the other hand, ignoring their dependence on the parameters (the TRBM) produces suboptimal inference; while including the dependence in the learning rules (the RTRBM) introduces 
backpropagation-through-time.
We show this in the following sections and argue why it is so in \sctn{conclusions}.

\FigTemporalEFHs

\subsection{The TRBM}
There are two training procedures for the graphical model of \subfig{RTRBMgraphicalmodel}.
The first employs the approximation we have made use of for the rEFH. 
In minimizing KL divergence between data, $p(\stateobsvs{t};\params)$, and model, $q(\stateobsvs{t}|\rcrnts{t-1};\params)$, the dependence $\rcrnts{t-1}$ on the parameters (\eqn{rcrntdstrb}) is ignored.
Applying the method of one-step contrastive divergence to the distribution in \eqn{RTRBMjointdstrb} then yields
\begin{equation}\label{eqn:TRBMparameterUpdates}
	\begin{split}
		\Delta\vishidwts &\propto \sum_{t=0}^T 
			\smplavg{
				q(\modelstates{t}|\rcrnts{t-1},\stateobsvs{t})
				p(\rcrnts{t-1},\stateobsvs{t})}
				{\Modelstates{t}\Stateobsvs{t}\tr
				-\smplavg{
					q(\updtmodelstates{t}|\rcrnts{t-1},\updtstateobsvs{t})
					q(\updtstateobsvs{t}|\rcrnts{t-1},\modelstates{t})}
					{\Updtmodelstates{t}\Updtstateobsvs{t}\tr
				}
			},\\
		\Delta\rcrnthidwts &\propto \sum_{t=0}^T 
			\smplavg{
				q(\modelstates{t}|\rcrnts{t-1},\stateobsvs{t})
				p(\rcrnts{t-1},\stateobsvs{t})}
				{\Modelstates{t}\Rcrnts{t-1}\tr
				-\smplavg{
					q(\updtmodelstates{t}|\rcrnts{t-1},\updtstateobsvs{t})
					q(\updtstateobsvs{t}|\rcrnts{t-1},\modelstates{t})}
					{\Updtmodelstates{t}\Rcrnts{t-1}\tr}				
			},\\
		\Delta\hidbiases &\propto \sum_{t=0}^T 
			\smplavg{
				q(\modelstates{t}|\rcrnts{t-1},\stateobsvs{t})
				p(\rcrnts{t-1},\stateobsvs{t})}
				{\Modelstates{t} 
				-\smplavg{
					q(\updtmodelstates{t}|\rcrnts{t-1},\updtstateobsvs{t})
					q(\updtstateobsvs{t}|\rcrnts{t-1},\modelstates{t})}
					{\Updtmodelstates{t}}
				},\\
		\Delta\visbiases &\propto \sum_{t=0}^T 
			\smplavg{
				q(\modelstates{t}|\rcrnts{t-1},\stateobsvs{t})
				p(\rcrnts{t-1},\stateobsvs{t})}
				{\Stateobsvs{t}
				-\smplavg{
					q(\updtstateobsvs{t}|\rcrnts{t-1},\modelstates{t})}
					{\Updtstateobsvs{t}}
				}.
	\end{split}
\end{equation}
We emphasize the subtle difference between \eqn{rEFHparameterUpdates} and \eqn{TRBMparameterUpdates}; in particular, no ``updated'' version of the posterior means, $\updtrcrnts{t-1}$, ever appears in the latter.

\subsection{The RTRBM}
If the dependence of the KL divergence on the parameters via the data distribution, introduced by \eqn{rcrntdstrb}, is \emph{not} ignored \citep[][]{Sutskever2009}, the learning rules accrue an extra term, which must be computed with backpropagation-through-time (BPTT).
Thus the RTRBM learning rules are
\begin{align*}
	\Delta\vishidwts
		&= (\Delta\vishidwts)_\text{TRBM} + \eta\sum_{t=0}^T \BPTTvec{t}\stateobsvs{t}\tr,
		&
	\Delta\rcrnthidwts
		&= (\Delta\rcrnthidwts)_\text{TRBM} + \eta\sum_{t=0}^T \BPTTvec{t}\rcrnts{t-1}\tr\\
	\Delta\hidbiases
		&= (\Delta\hidbiases)_\text{TRBM} + \eta \sum_{t=0}^T \BPTTvec{t},
		&
	\Delta\visbiases
		&= (\Delta\visbiases)_\text{TRBM};\\
	\BPTTvec{t}
		&= \transitionjacobian{t}\tr\rcrnthidwts\tr \bigg(
			\colgradient{\log q(\stateobsvs{t+1}|\rcrnts{t};\params)}{\hidbiases} 
			+ \BPTTvec{t+1}\bigg).
\end{align*}
where the last line defines the backwards recursion for BPTT (for a derivation, see \app{BPTT}), $\eta$ is the learning rate, and $\transitionjacobian{t}$ is the Jacobian of the nonlinearity (e.g., logistic function for Bernoulli units).\footnote{Of course, the nonlinearity acts element-wise, so the Jacobian is diagonal, and in practice is implemented as a Hadamard product rather than a matrix multiplication.}


\section{Experiments}\label{sec:experiments}

We compare performance of the rEFH, TRBM, and RTRBM on two different kinds of data:\ a linear dynamical system (LDS) that is observed via a ``probabilistic population code'' (PPC) \citep{Ma2006}; and the bouncing-ball data used by \citet{Sutskever2007} and \citet{Sutskever2009}.
For the LDS, we have reported previously a subset of the results for the rEFH \citep{Makin2015b}.
Here we compare the filtering performance of the rEFH with that of the TRBM and RTRBM on these data, and investigate the failure modes of the latter, especially in connection with the order of the dynamics that can be learned.
For the bouncing-ball data, \citet{Sutskever2013} characterized the performance of the TRBM and RTRBM in terms of prediction of the next frame.
Here, we characterize the rEFH's performance in the same way on the same data.

\subsection{Details of the training procedures}
With the exception of the rEFHs trained on LDSs (\sctn{LDSwithPPCs}), all models and experiments used the same training scheme, which was matched to that reported by \citet{Sutskever2013}.
Weight changes were made according to 25-step contrastive-divergence with a fixed momentum of 0.9 and a learning rate declining linearly from 1/1500 (LDS) or 1/100 (bouncing balls)\footnote{The differing learning rates reflect the size of the inputs:\ the LDS data are Poisson distributed with maximum mean firing rates of about 10, whereas the bouncing-ball ``pixels'' are Bernoulli distributed.}, on ``minibatches'' consisting of 100 time steps of a single trajectory.
New batches of 400 trajectories (i.e., 400*100 = 40000 sample vectors) were created every 5 epochs, for 250 epochs.
These models were initialized with 30 batches of static pre-training (i.e., setting recurrent activities to zero) at CD-5.

The rEFHs of \sctn{LDSwithPPCs}, on the other hand, could be trained much more quickly.
No static pre-training was required, only CD-1 was used throughout, and training terminated after only 90 epochs.
Minibatches consisted of 40 vectors, each corresponding to the same time step of 40 different trajectories.\footnote{This minibatch scheme is precluded for the RTRBM by the use of BPTT, which requires many steps of the \emph{same} trajectory.}
Batches, renewed every 5 epochs, consisted of 1000 time steps (i.e., 40*1000 = 40000 sample vectors).
Learning rates declined \emph{exponentially}, according to $1.1^{-\text{epoch}}$, from 1/500 (Poisson-Bernoulli weights), 1/50 (Bernoulli-Bernoulli weights) or 1/120 (biases).
Momentum started at about 0.5 and exponentially approached 0.98 ($\rho = 0.98 - 0.5*1.1^{-\text{epoch}}$).
A small weight decay of 0.001 was also used.

\subsection{Linear dynamical system with PPC observations}\label{sec:LDSwithPPCs}

We aimed to quantify model performance precisely, so we constructed a generative model for the data (consistent with \subfig{HMMaugmented}) for which the posterior distribution can be computed in closed form---despite nonlinear, non-Gaussian emissions.
The ``backbone'' of the model is a second-order linear dynamical system; that is, position and velocity depend on previous position and velocity.
The observations at each time step are a product over fifteen conditionally independent Poisson ``neurons'' with fixed-width Gaussian-shaped tuning to the position---a ``probabilistic population code'' \citep{Ma2006}.
This could be thought of as a crude model of sensory responses to a moving object.
Since the tuning curves evenly tile the space, the posterior distribution over the hidden dynamical state is in fact (approximately) a normal distribution \citep{Ma2006}.
This makes the filtering distribution (the RHS of \eqn{posteriorGivenHiddensIsFilteringDistribution}) computable in closed form---the Kalman filter (KF)---where, in place of the observation and the emission variance, respectively, one uses the center of mass of the population code and the tuning-curve width scaled by the total number of spikes \citep{Ma2006}.
This model can be learned with an EM algorithm \citep{Ghahramani1996a}, which allows us to benchmark the rEFH, TRBM, and RTRBM against easily interpreted quantities.
In particular, we can consider Kalman filters applied to models learned by EM under the assumption of first- (\KFone) or second-order (\KFtwo) dynamics.
We also considered a zeroth-order model (\KFnaught), which assumes no dynamics at all; and a KF that has access to the parameters of the true dynamical system (\KFopt---no learning required).
We chose these parameters so that any first-order approximation will be poor, allowing us to distinguish these cases from each other.

Fresh data were generated for testing:\ for each model the posterior mean, $\xpct{q}{\Modelstates{t}|\augobsvs{t}}$, was computed at each time step.
From each vector of hidden activities, the position of the underlying ``stimulus'' was decoded (using the technique we have introduced for similar, static models; see \citealt{Makin2013b} or \app{experimentdetails}) and used to compute an error with respect to the actual position.
We did the same using the posterior mean from Kalman filters learned using \KFone, \KFtwo; the zeroth-order model; and the ``optimal'' (no-learning) model.
For each, we report the mean, across time steps (1000) and trajectories (40), of the square of these errors (MSE).
In order to characterize susceptibility of training to local minima, 20 of each of the EM-based models (\KFone, \KFtwo) were trained from scratch.
Likewise, 20 of each of the rEFH, TRBM, and RTRBM were trained from scratch for each of 20 different hidden-layer sizes.
The distribution of the resulting MSEs are shown in the box-and-whisker plot in \fig{ErrorStatsVsNumHiddens}.
Also indicated with lines are the performances of the \emph{best} of each of the \KFone\ and \KFtwo\ models, and the performances of the models requiring no training, \KFnaught\ and \KFopt.
(Since a single set of testing data was used for all models, there is no variation in performance---or, therefore, box plots---for the no-training models.)

Evidently, the TRBM learns only first-order dynamics, as suggested by \citet{Sutskever2013}, who reported that it tends to generate data that move in a random walk.
The best RTRBMs approach the performance of the best second-order Kalman filter (\KFtwo, the order of the true underlying dynamics).
The rEFHs are on average slightly worse, but given at least 195 hidden units, the best are very close to the best RTRBMs, which do not appear to improve after about 165 hidden units.
The rEFHs are certainly able to learn second-order dynamics, unlike the TRBM; and they do not require BPTT, unlike the RTRBM.
In fine, the performance deficit of the rEFH relative to the RTRBM appears not to be insuperable, as with the TRBM, but rather to be remediable---at least on these data---with an increase in hidden units.
This cost may be offset to some extent by the savings accrued from not needing BPTT; we explore this below on a more difficult task, the bouncing-ball data set.
Finally, we note that only outliers among the \KFtwo\ models actually achieve performance superior to that of the median rEFH or RTRBM, as long as these neural networks have at least 60 hidden units.

\FigErrorStatsVsNumHiddens


\subsection{Bouncing balls}
Each data vector consists of a 30$\times$30 image patch containing three ``balls'' (circles) that bounce off each other and the walls with complete energy conservation.
The nonlinear dynamics greatly complicate computation of the posterior distribution over the hidden state---the positions and velocities of the balls---so we compute from the hidden units not the optimal posterior means, but the best predictions of the next frame, as proposed by \citet{Sutskever2013}.
We note first of all that the \emph{current frame} predicts the next frame with a per-pixel, mean square error (MSE) of 0.015.

Our implementation of the TRBM, with 400 hidden units, achieves a MSE of 0.046, close to the 0.04 reported by \citet{Sutskever2013}.
Interestingly, this is higher (worse) than the current-frame prediction, which suggests that this TRBM does not learn any dynamical model at all.
In fact, this is consistent with the results above:
The TRBM can learn only first-order dynamics; and for the bouncing balls, the optimal first-order model is the zeroth-order model, since the velocities are indeed constant (until collisions), and the \emph{average}, over all trajectories, of the next position is the current position (velocities are on average zero).
That the TRBM fails even to learn even the zeroth-order model (the current-frame predictions) suggests, perhaps, a limitation imposed by the nonlinearities (collisions) in the dynamics.

Our implementation of the RTRBM, with 400 hidden units, achieves a MSE of 0.008, again close to the results reported by \citet{Sutskever2013}, in this case 0.007.
The rEFH with 400 hidden units achieves a per-pixel MSE of 0.014:\ superior to the TRBM, as well as the zeroth- and first-order predictions, but inferior to the RTRBM.
However, training time is also much shorter (see \tbl{bbresults}), since BPTT is not required.\footnote{The comparison is between our own implementations of the RTRBM and the rEFH.
Our highly vectorized, GPU-optimized {\sc Matlab} implementations take on the order of hours; on the same machine, Sutskever's original {\sc python} implementation of the RTRBM takes a few days.
All of our code is publicly available at \tt{https://github.com/jgmakin}.}
Increasing the number of hidden units to 625 increases the rEFH's training time to that of the RTRBM, about 2.5 hours, and brings its MSE closer, down to 0.010 MSE.
Allowing a little under five hours accommodates 1000 hidden units in the rEFH and approximately matches the RTRBM's MSE at 0.008.

\begin{table}[ht]
\begin{minipage}{\textwidth}
\begin{center}
\begin{tabular}{c|c|c|c}
model & $\#$ hidden & MSE & training time (hrs)\\
\hline
\hline
0$^\text{th}$-order & --- & 0.015 & ---\\
TRBM	& 400	& 0.046 	& 1.43\\
TRBM	& 625	& 0.040		& 2.53\\
RTRBM 	& 400 	& 0.008		& 2.56\\
rEFH 	& 400 	& 0.014 	& 1.42\\
rEFH 	& 625 	& 0.010		& 2.56\\
rEFH 	& 1000 	& 0.008 	& 4.76\\
\end{tabular}
\caption{For the bouncing-ball data, a comparison of next-frame per-pixel MSE, and training times.}
\label{tbl:bbresults}
\end{center}
\end{minipage}
\end{table}

\subsection{Generating trajectories}\label{sec:generation}
The rEFH enjoys a different advantage over the (R)TRBM:\ generating sequences is orders of magnitude less costly.
Because the rEFH defines a distribution over the previous hidden units, generation in the reverse direction is particularly simple:\ from each hidden vector, a ``previous'' hidden vector as well as the current observation can be drawn immediately, and the process iterated.
In the (R)TRBM, by contrast, generating sequences (in either direction) requires at each time step repeated Gibbs sampling of the hidden and visible units, conditioned on the previous hidden means, before the draws converge to the stationary distribution.
In practice, this appears to be about 50 steps---i.e., 50 upward and 50 downward passes \citep{Sutskever2013}.
Examples of bouncing-ball trajectories generated by the rEFH can be found in the supplemental material.

\section{Conclusions}\label{sec:conclusions}
The ``recurrent exponential-family harmonium'' (rEFH) generalizes EFHs to time-series data.
Its central motivation is to provide optimal inference to the hidden state of ``the world'' that produces those data, even when the world's true generative model is intractable to Bayesian inversion, hard to express in closed-form equations, or even unknown.
Thus, in contrast to directed graphs like dynamic Bayes Nets \citep{Murphy2002}, where one attempts to formulate the true generative model and then derive an inference algorithm for it, we relaxed the assumption that the generative model of the world and the rEFH correspond, in favor of a weaker one about identifiability: 
The rEFH's and the world's marginal distributions over observations must be expressible in terms of the same identifiable finite mixture model (see \app{technicalconditions} for technical details).
Consequently, the sufficient statistics for the true hidden state are not derived in closed form (like, e.g., the posterior cumulants recursively computed in a Kalman filter), but are nevertheless guaranteed to be encoded in the hidden units of the trained rEFH.
The idea behind the proof of this guarantee is simple:\ good density estimators make their hidden units---or rather, the sufficient statistics for them---good summaries of their input data.
When such summaries are also included among the inputs---at the next time step (\fig{tikzHMMs})---the succeeding hidden vector is forced to summarize both the new inputs and the old summaries.
This recursive procedure builds up a master summary that takes into account all relevant past information. 

The other conspicuous consideration regarding the choice of directed and undirected (or at least EFH-like) models is the trade-off between inference and generation.
In directed models, generation is easy, but inference---in all but a few special cases---is hard, and approximate techniques are necessary; whereas in the EFH-like undirected models, inference (to its own hidden states) is easy, but generation is hard, because it requires Gibbs sampling.
But in the rEFH, generating sample trajectories---in reverse time order---is easy, requiring only a single pass through the model.
This is not the case in the alternative architecture proposed by \citet{Sutskever2013}, the TRBM and RTRBM (\subfig{RTRBMgraphicalmodel}), where prolonged Gibbs sampling is required.

In practice, some of the assumptions of the proof are mildly violated:\ e.g., that the model has become a perfect model for the training data.
To show that the proof is robust against such violations, we tested the model experimentally on data from a generative model that has been contrived to yield closed-form solutions to the inference problem, a second-order system whose dynamics are, and whose emissions can be re-formulated to be, linear-Gaussian.
This enabled us to demonstrate conclusively that the rEFH learns second-order dynamics from reports of position only---even though its recurrent connections span only one time step.

But the most interesting approximation is the use of a sub-optimal learning rule.
During learning, changes to the model's parameters ramify into the data, since they consist in part of the model's own hidden-unit means (\eqn{rcrntdstrb}).
We ignored this, which amounts to assuming that whatever improvements to the model are effected by a parameter update are not (entirely) undone by the ensuing changes to the data.
This obviates the need for backprop through time (BPTT).
In Sutskever's architecture (\subfig{RTRBMgraphicalmodel}), the same approximation turns the RTRBM into a TRBM, and loses the ability to learn dynamics beyond first-order, at least without increasing the span of its recurrent connections.
In contrast, the only advantage the RTRBM appears to enjoy over the rEFH is more economical use of hidden units (\fig{ErrorStatsVsNumHiddens}, \tbl{bbresults}), which comes at the price of slower training times (BTPP being required), and the inability to generate trajectories in one pass.
Some of the rEFH's training-time advantage can be bartered for performance by increasing the size of its hidden layer ---although to eliminate completely the performance gap on Sutskever's bouncing-ball data set ultimately requires about a factor of 2 \emph{more} time for the rEFH (\tbl{bbresults}).
But for data with trajectories longer than 100 time steps, the advantage of neglecting BPTT is even greater, and the relative merits of the rEFH increase.

Why can BPTT be neglected for the rEFH but not the (R)TRBM?
The answer does not follow obviously from application of BPTT to the rEFH architecture, which yields non-zero terms.
We consider this the most important theoretical question for future work.
Second, empirically, the accumulation of master summaries described above is ultimately limited, obviously, by the capacity of the hidden layer, so not all past history can be ``summarized'' in the way just described.
(This is consistent with the proof because the capacity limits would likewise limit the generative fidelity of the model to the data, compromising one of the proof's requirements.)
Finally, it also seems unlikely that the model, without modification, could learn ``pathologically'' long dependencies.
Thus, although the toy data sets we picked allowed us to benchmark the models, it remains to determine how the model performs on more challenging time-series data.



\acks{Funding was provided by DARPA (N66001-10-C-2010, W911NF-14-2-0043), NIH Vision Training Grant (5T32EY007120-22), the UCSF Wheeler Center for the Neurobiology of Addiction, and (BKD) the National Science Foundation Graduate Research Fellowship under Grant No.\ 1144247.
Some of the EFHs were trained using Tesla K40 GPUs, the generous donation of the Nvidia Corporation.}


\appendix
\rvshortsequencemacroize{stateobsv}
\rvshortsequencemacroize{modelstate}
\rvshortsequencemacroize{worldstate}
\section{Information retention and identifiability}\label{sec:technicalconditions}
The essential result used to prove optimal inference in the rEFH comes from \citet{Makin2015a}, who showed that for a certain class of latent-variable density estimators, depicted in \fig{tikzWorldModel}, the following implication holds:
\begin{equation}\label{eqn:DensityEstimationImpliesInformationRetention}
	q(\augobsvs{t}) = p(\augobsvs{t}) 
		\implies 
	p(\worldstatestillnow|\augobsvs{t}) = p(\worldstatestillnow|\ssformodelstates{t})
		\implies 
	p(\worldstates{t}|\augobsvs{t}) = p(\worldstates{t}|\ssformodelstates{t}).
\end{equation}
Time-dependence is not discussed in that paper, but we have subscripted the variables in \eqn{DensityEstimationImpliesInformationRetention} with $t$ to facilitate our application of that result to the models of the present paper.
The equation can be read as saying that a match between the model and data marginals (the left-hand equation) implies that the posterior distribution over all true hidden states up to time $t$, $\worldstatestillnow$, is the same whether conditioning on the observed data at that time step, $\augobsvs{t}$, or on any set of sufficient statistics, $\ssformodelstates{t}$, for the \emph{model}'s hidden states, $\Modelstates{t}$.
For the EFH, the sufficient statistics consist of the vector of hidden-unit means.
(We have used ``U'' rather than the usual ``T'' for sufficient statistics because the latter is used for final time.)
The final equation says that the same applies (\emph{a fortiori}) to the posterior distribution over just the \emph{current} true hidden state, $\Worldstates{t}$.
More colloquially, we can say that training the model to be a good density estimator guarantees that no information about the true hidden state will be lost when making a (deterministic) ``upward'' pass through the model, from the visible to hidden units.

In the experiments, and in the proof in the main text, the data vector $\augobsvs{t}$ consists of the current observations, concatenated with the previous hidden-unit means.
Our exposition of identifiability is greatly facilitated, however, by working with \emph{samples}, rather than means, from those hidden units:
The EFH, after all, produces Bernoulli samples, not real numbers, on its ``left side'' (see \subfig{EFHinrEFH}).
We justify this approximation by noting that, for sufficiently long hidden vectors, little information will be lost even from a \emph{sample} from the vector of hidden means.
Indeed, when sample rather than mean vectors were used in experiments like those in the main text, there was no noticeable performance difference \citep{Makin2015b}.
Thus we replace \eqn{augmentedData} in the main text with
\begin{equation}\label{eqn:augobsvsDefnInPractice}
	\augobsvs{t} := \begin{bmatrix}\modelstates{t-1} & \stateobsvs{t}\end{bmatrix}.
\end{equation}

Previously \citep{Makin2015a}, we provided sufficient conditions for membership in the class of models for which \eqn{DensityEstimationImpliesInformationRetention} holds.
Here we consider those technical conditions with respect to the dynamical models discussed in the present paper.
We discuss both the specific model on which the experiments (especially those of \sctn{LDSwithPPCs}) were performed, and dynamical models more generally.
Although we argue below that satisfaction of the conditions is plausible, we suspect that in practice some of them are mildly violated; we indicate where below.
The experiments show, what the proof does not, that the guarantees are robust to such violations, in the sense that mild departures from the assumptions lead only to mild departures from the conclusion, optimal inference.

\FigTikzWorldModel

\paragraph{Condition 1.}
The first technical condition is that the observations $\Augobsvs{t}$ be the same type of random variable in both the ``world'' ($p$, the ``data distribution,'' in Hinton's terminology) and the ``model'' ($q$, ``model distribution'').
More precisely, the conditional distributions $p(\augobsvs{t}|\worldstatestillnow)$ and $q(\augobsvs{t}|\modelstates{t})$ (see again \fig{tikzWorldModel}) must both be expressible in terms of some common conditional distribution $f(\augobsvs{t}|\proxcauses{t})$, with $\proxcauses{t} = \worldstateSSfunc{\worldstatestillnow}$ in the world, and $\proxcauses{t} = \modelstateSSfunc{\modelstates{t}}$ in the model, for some functions $\worldstateSSfunc{\cdot}$ and $\modelstateSSfunc{\cdot}$.
For example, if the observations were truly ($p$) generated by products over Poisson distributions, the model ($q$) observations must be expressible as products over Poissons, as well.

In real and artificial neural networks, this condition is not particularly restrictive:\ it says that neurons must respond with the same \emph{kind} of statistics---Poisson, Bernoulli, Gaussian, etc.---whether driven bottom-up or top-down.
In the experiments in the main text, the hidden units were Bernoulli random variables while the ``sensory'' units were Poisson.
Using samples (rather than means) for the recurrent units, then, makes the function $f$ a product over Bernoulli distributions and Poisson distributions.
What about $\modelstateSSfunc{\modelstates{t}}$ and $\worldstateSSfunc{\worldstatestillnow}$?
For the model, $\modelstateSSfunc{\modelstates{t}}$ was a ``downward'' (hidden-to-visible) pass through the weight matrix, followed by the element-wise nonlinearities:\ the logistic function for Bernoulli units, the exponential function for the Poisson units.
For the ``world,'' $\proxcauses{t} = \worldstateSSfunc{\worldstatestillnow}$ was more complicated.
The ``right-hand'' subvector (cf.\ \eqn{augobsvsDefnInPractice}), for the Poisson units, consisted of the embedding of the current hidden state $\worldstates{t}$ into (Gaussian) tuning curves.
The ``left-hand'' subvector, for the Bernoulli units, was created with the recursion defined by \eqn{rcrntdstrb} (see also \fig{tikzHMMs} in the main text).
For simplicity in the following discussion, we substitute samples $\modelstates{t-1}$ for the mean vector $\xpct{q}{\Modelstates{t-1}|\augobsvs{t-1};\params}$.

\paragraph{Condition 2.}
The second condition is the antecedent proposition in \eqn{DensityEstimationImpliesInformationRetention}, sc., that the model density match the data density.
Training the EFH as a density estimator---that is, changing its parameter so as to descend the gradient of a function with $q(\augobsvs{t}) = p(\augobsvs{t})$ at its minimum---is supposed to guarantee this.
It must be noted, however, that the proof in our previous work \citep{Makin2015a} requires strict equality, whereas the use of $n$-step contrastive divergence with small $n$ \citep[which is believed to produce suboptimal density estimators; see][]{Hinton2010} and the finite number of training samples suggests that the equality need only be approximate in practice.

\paragraph{Condition 3.}
The third condition is that the number of hidden-variable states be finite.
Although the data in the experiments were in fact generated with real-valued hidden states, $\Worldstates{t}$, we rely on the fact that only a finite number of such states was ever generated.
This condition is in fact presupposed by the antecedent proposition, $q(\augobsvs{t}) = p(\augobsvs{t})$:\ since the EFH has a finite, albeit enormous ($2^\text{number of hidden units}$), number of hidden states, the model distribution cannot be guaranteed to match the data distribution unless the latter likewise has a finite (albeit possibly enormous) number of hidden states.

\paragraph{Condition 4.}
The final condition is the most complicated.
To begin with, we re-write the data and model distributions in terms of $\Proxcauses{t}$:\ see \fig{tikzWorldModelWithProxcauses}.
Since the number of hidden states is finite, these can be thought of as finite mixture models.
The required condition is that the mixture of the conditional distributions, $f(\augobsvs{t}|\proxcauses{t})$, be ``identifiable,'' which is to say that the map from the parameters of the mixture to the marginal $p(\augobsvs{t})$ be injective (up to relabeling of the hidden units).
In particular, in identifiable mixture models, $p(\augobsvs{t}) = q(\augobsvs{t}) \implies p(\proxcauses{t}) = q(\proxcauses{t})$, which is the key to proving \eqn{DensityEstimationImpliesInformationRetention}.

\FigTikzWorldModelWithProxcauses

Identifiability in this sense is the rule rather than the exception \citep{Titterington1985}.
Nevertheless, several exceptions are notable, among which is the RBM, a mixture of products of Bernoulli distributions \citep{Titterington1985}.
On the other hand, a recent result \citep{Allman2009} restores identifiability to many of these exceptions by relaxing the definition very mildly.
In particular, \citet{Allman2009} consider ``generically'' identifiabe mixture models, that is to say, those that are unidentifiable only on sets of measure zero.
Their results imply, among other things, that the RBM is generically identifiable as long as there are more than twice as many visible as hidden units.
For the bouncing-ball data, then, the RBM can have up to 899 hidden units---each image being itself 900 pixels, yielding a total of 1799 visible (image plus recurrent) units.

For the EFH in our linear-dynamical-system experiments, application of the results of Allman is slightly more involved, since the conditional distribution $f(\augobsvs{t}|\proxcauses{t})$ is a product over both Bernoullis and Poissons.
The relevant result is Theorem 4 in that paper, which supplies identifiability conditions for finite mixtures of finite-measure products.
Consider in particular an EFH with:\ $N$ hidden units, each with cardinality (number of possible states) $n$; $M$ non-recurrent units, each with cardinality $m$; and therefore $N + M$ visible units, with the $\jth$ hidden unit having cardinality $k_j$.
Distributions defined by this model are generically identifiable if \citep{Allman2009}
\begin{equation}\label{eqn:AllmanTheoremFour}
	\min(r,\kappa_1) + \min(r,\kappa_2) + \min(r,\kappa_3) \geq 2r + 2,
\end{equation}
where we have defined $r = n^N$, the number of hidden \emph{states}; integers $\kappa_i := \prod_{j\in S_i}k_j$; and sets $S_1,S_2,S_3$, which constitute a disjoint, exhaustive partition of the visible units, $\augobsvs{t}$.

Assuming $r > 2$, the optimal strategy for partitioning the visible units is to try to have $\kappa_1 = \kappa_2 = r$, allowing $\kappa_3$ to be whatever remains, since $\min(r,\kappa_3)$ is guaranteed to be at least 2 in any case.
Thus, $S_3$ should contain just one visible unit, and $S_1$ and $S_2$ should each contain half the remainder, divided so that $\kappa_2$ and $\kappa_2$ are (roughly) equal.
Under this scheme, we have $\kappa_2 = \kappa_3 \approx n^{N/2} m^{M/2}$.
(We assume that $N$ and $M$ are much greater than 1, allowing us to ignore the one unit assigned to $S_3$.)
Requiring that this be greater than $r = n^N$ is tantamount to requiring that
\begin{equation}\label{eqn:requiredCardinalityOfPoissons}
	n^N < m^M.
\end{equation}

What about our Poisson-Bernoulli EFH?
Technically, the theorem applies only to finite-measure products, whereas the Poisson units can take on an infinite number of states.
On the other hand, the conditional distribution of each unit can be approximated arbitrarily well by a sufficiently long, but truncated, version of the Poisson distribution.
Under such an approximation, \eqn{AllmanTheoremFour} is satisfied, each EFH is generically identifiable, and the perfectly trained rEFH is guaranteed (with probability 1) to provide optimal inference about the true hidden causes of the linear dynamical system---i.e., \eqn{DensityEstimationImpliesInformationRetention} holds.

On the other hand, identification of such a model would require an impractically large number of training samples.
For any finite number of samples, almost all of the infinite number of states that a Poisson random variable can take on will have never be observed; thus, their probability will be indistinguishable from zero.
In our experiments, we tested rEFHs with $M=15$ Poisson units, and $N$ between 15 and 300 hidden/recurrent Bernoulli units, having cardinality 2.
If we choose $\kappa_2$ as above, satisfying \eqn{requiredCardinalityOfPoissons} for, say, $N=150$, requires that $m=1000$, whereas the largest Poisson mean was about 10, making a nonzero count (in the number of samples generated for the experiments) essentially zero for any integer greater than about 30.

We have shown, then, that the model used in the linear-dynamical-system experiments is formally generically identifiable, in the sense of \citet{Allman2009}.
Nevertheless, in practice, the limited number of training samples will limit this identifiability, due to both condition 2 and condition 4.
The fact that this does not prevent the trained model from providing good inference to the true hidden state suggests either that the training procedure is robust against this type of failure---that mild departures from identifiability result in mild departures from optimal inference---or that identifiability is not in fact required for \eqn{DensityEstimationImpliesInformationRetention} in this model.
Indeed, it is not in general \citep{Makin2015a}, although no other set of sufficient conditions has been adduced.
In either case, \eqn{requiredCardinalityOfPoissons} suggests how to modify rEFHs in order to guarantee identifiability.
Generally speaking, the non-recurrent units must be either more numerous, or able to take on more states, than the recurrent/hidden units.


\section{Derivation of the update rules for the RTRBM}\label{sec:BPTT}
Since our update equations differ slightly from those given by \citet{Sutskever2009}, and no explicit derivation is given in that work, we derive the training algorithm for the RTRBM here.
(We use transposes---denoted $\tr$, since ``T'' is used for final time---to indicate the shape of our matrix derivatives.
For now, $\params$ is a generic vector of parameters.)
First, we consider a generic, deterministic, recurrent neural network (RNN) with inputs $\Stateobsvsalltime \sim p(\stateobsvsalltime)$ and ``hidden''/output-unit activities $\modelpostmeans{t}$ recursively computed as
\begin{equation}\label{eqn:genericrecurrency}
	\modelpostmeans{t} = f(\modelpostmeans{t-1},\stateobsvs{t},\params).
\end{equation}
We leave the loss function undetermined for now, but let it depend on the output at every time step, as well as directly on the parameters:
\begin{equation}\label{eqn:genericloss}
	L(\modelpostmeans{0}(\params),\ldots,\modelpostmeans{T-1}(\params),\params).
\end{equation}
To compute parameter changes, we take the (total) derivative of the average (under the data distribution) loss function: 
\begin{equation}\label{eqn:lossderivativeA}
	\begin{split}
		\ttlderiv{}{\params}\smplavg{p(\stateobsvsalltime)}{L}
			&= \smplavgg{p(\stateobsvsalltime)}{\jacobian{L}{\params} 
				+ \sum_{t=0}^T \jacobian{L}{\modelpostmeans{t}}\ttlderiv{\modelpostmeans{t}}{\params}}\\
			&= \smplavgg{p(\stateobsvsalltime)}{\jacobian{L}{\params} + \sum_{t=0}^T \ttlderiv{L}{\modelpostmeans{t}}\jacobian{\modelpostmeans{t}}{\params}}.
	\end{split}
\end{equation}
The first equality is just standard calculus; the second follows because $\params$ only occurs in the final terms in the chains of derivatives.
Thus, one can either compute the total effect of $\params$ on each $\modelpostmeans{t}$, and then compute the direct effects of the latter on $L$; or one can compute the \emph{direct} effect of $\params$ on each $\modelpostmeans{t}$, and then compute the \emph{total} effect of the latter on $L$.

Because of \eqn{genericrecurrency}, the effect on $L$ of $\modelpostmeans{t}$, the current hidden state, is the sum of its direct effect and the effects caused by its influence on \emph{future} hidden states, $\modelpostmeans{t+k}, k > 0$. 
This can be expressed recursively as
\begin{equation}\label{eqn:backpropthroughtime}
	\ttlderiv{L}{\modelpostmeans{t}} 
		= \jacobian{L}{\modelpostmeans{t}} 
		+ \ttlderiv{L}{\modelpostmeans{t+1}}\jacobian{\modelpostmeans{t+1}}{\modelpostmeans{t}}.
\end{equation}
The recursion is initialized at $\dfrntl{L}/\dfrntl{\modelpostmeans{T}\tr} = \partial L/\partial\modelpostmeans{T}\tr$ and run backwards in time.
The backprop-through-time (BPTT) algorithm consists of \eqns{lossderivativeA}{backpropthroughtime}.
Applying it to any particular model requires only working out four partial derivatives, 
$\partial L/\partial\modelpostmeans{t}\tr$,
$\partial L/\partial\params\tr$, 
$\partial \modelpostmeans{t+1}/\partial\modelpostmeans{t}\tr$, and
$\partial \modelpostmeans{t}/\partial\params$, 
for a particular choice of recurrent function and loss function.

Now we show why BPTT arises in the context of training an RTRBM.
For the RTRBM, the average loss function is the KL divergence between data and model distributions, 
$p(\stateobsvsalltime)$ and $q(\stateobsvsalltime;\params)$, respectively.
The latter, however, can be re-expressed using \eqn{rcrntdstrb} and the graph for the RTRBM, \subfig{RTRBMgraphicalmodel}:
\begin{equation}\nonumber
	\begin{split}
		q(\stateobsvsalltime;\params)
			&= \prod_{t=0}^T \int_{\rcrnts{t-1}}
				q(\stateobsvs{t}|\rcrnts{t-1};\params)
				p(\rcrnts{t-1}|\stateobsvstillprev;\params)\dfrntl{\rcrnts{t-1}}\\
			&= \prod_{t=0}^T \int_{\rcrnts{t-1}}
				q(\stateobsvs{t}|\rcrnts{t-1};\params)
				\delta(\rcrnts{t-1} - \modelpostmeans{t-1}(\stateobsvstillprev,\params))\dfrntl{\rcrnts{t-1}}\\
			&= \prod_{t=0}^T q(\stateobsvs{t}|\modelpostmeans{t-1}(\stateobsvstillprev,\params);\params).
	\end{split}
\end{equation}
The first line follows from the graph; the second from \eqn{rcrntdstrb}.
Here, $\modelpostmeans{t}$ is the posterior mean, and therefore computable with a deterministic ``upward'' pass through the model,
\begin{equation}\label{eqn:RTRBMrecurrency}
	\modelpostmeans{t} = f(\rcrnthidwts\modelpostmeans{t-1} + \vishidwts\stateobsvs{t} + \hidbiases),
\end{equation}
where the function $f$, acting element-wise on its vector argument, is determined by the exponential family 
to which the posterior distribution belongs (logistic for Bernoulli, exponential for Poisson, etc.).
Now writing the KL divergence in terms of our new expression for the model distribution, we find that
\begin{equation}\nonumber
	\begin{split}
		\ttlderiv{}{\params} \KLop{p(\stateobsvsalltime)}{q(\stateobsvsalltime;\params)}
			&= -\ttlderiv{}{\params} \smplavg{p(\stateobsvsalltime)}{\log q(\stateobsvsalltime;\params)}\\
			& = -\ttlderiv{}{\params} \smplavgg{p(\stateobsvsalltime)}{\log\prod_{t=0}^T 
				q(\stateobsvs{t}|\modelpostmeans{t-1}(\stateobsvstillprev,\params);\params)}\\
			& = -\ttlderiv{}{\params} \smplavgg{p(\stateobsvsalltime)}{\sum_{t=0}^T
				\log q(\stateobsvs{t}|\modelpostmeans{t-1}(\stateobsvstillprev,\params);\params)}\\
			& = \ttlderiv{}{\params} \smplavgg{p(\stateobsvsalltime)}{\sum_{t=0}^{T}
				L_t(\modelpostmeans{t-1}(\params),\params)}
			  = \ttlderiv{}{\params}\smplavg{p(\stateobsvsalltime)}{L},
	\end{split}	
\end{equation}
where we have defined
\begin{equation}\label{eqn:RTRBMloss}
	\begin{split}
		L_t(\modelpostmeans{t-1}(\params),\params) 
			&:= -\log q(\stateobsvs{t}|\modelpostmeans{t-1};\params),\\
		L(\modelpostmeans{0}(\params),\ldots,\modelpostmeans{T-1}(\params))
			&:= \sum_{t=0}^{T} L_t(\modelpostmeans{t-1}(\params),\params).
	\end{split}
\end{equation}
So improving (decreasing) this loss is equivalent to making the RTRBM a good generative model for the data.

Comparing \eqns{RTRBMrecurrency}{RTRBMloss} with \eqns{genericrecurrency}{genericloss}, we see that the RTRBM has the form of a recurrent neural network, with the posterior means playing the role of the hidden/output units, and the loss function depending on the parameters directly, as well as indirectly through its dependence on these ``outputs.''
Applying BPTT, \eqns{lossderivativeA}{backpropthroughtime}, to this recurrent function and this loss means working out $\partial L/\partial\modelpostmeans{t}\tr$,
$\partial L/\partial\params\tr$, 
$\partial \modelpostmeans{t+1}/\partial\modelpostmeans{t}\tr$, and
$\partial \modelpostmeans{t}/\partial\params\tr$.

We start with the partial derivatives of the loss.
First, from \eqn{RTRBMloss}, they decouple over time:
\begin{equation}\label{eqn:derlossderparams}
	\rowgradient{L}{\params}  = \sum_{t=0}^T \rowgradient{L_t}{\params},\qquad
	\rowgradient{L}{\modelpostmeans{t}}  = \rowgradient{L_{t+1}}{\modelpostmeans{t}}.
\end{equation}
Next, we recall that the distribution in \eqn{RTRBMloss} is the marginal over the visible units of an EFH (or RBM).
(Recall from \eqn{RTRBMjointdstrb} in the main text that each RBM or EFH in the RTRBM is defined via a conditional distribution, $q(\stateobsvs{t+1},\modelstates{t+1}|\modelpostmeans{t};\params)$.)
Hence, the partial derivatives with respect to the parameters, $\partial L_t/\partial \rcrnthidwts$, $\partial L_t/\partial\vishidwts$, and $\partial L_t/\partial \hidbiases$, are precisely the terms that are normally computed during unsupervised learning of the inputs, $\stateobsvs{t+1}$, in a standard EFH.
In the RTRBM, these gradients can be approximated with contrastive divergence (CD).

The gradient of the loss with respect to the hiddens/outputs---the quantity in the second equality in \eqn{derlossderparams}---turns out also to be computable with the standard unsupervised method for EFHs.
Beginning with the well known formula for the gradient of the log of the Boltzmann distribution,
\begin{equation}\label{eqn:derlossderhids}
	\begin{split}
		\rowgradient{L_{t+1}}{\modelpostmeans{t}}
			&= \xpctt{q(\modelstates{t+1},\stateobsvs{t+1}|\modelpostmeans{t})}
					{\jacobian{H}{\modelpostmeans{t}}\bigg|\modelpostmeans{t}}
				-\xpctt{q(\modelstates{t+1}|\stateobsvs{t+1},\modelpostmeans{t})}
					{\jacobian{H}{\modelpostmeans{t}}\bigg|\stateobsvs{t+1},\modelpostmeans{t}},\\
		H 
			&= \Modelstates{t+1}\tr \vishidwts \Stateobsvs{t+1} 
				+ \visbiases\tr\Stateobsvs{t+1} 
				+ (\rcrnthidwts\modelpostmeans{t} + \hidbiases)\tr\Modelstates{t+1}.\\
		\implies \jacobian{H}{\modelpostmeans{t}} 
			&= \Modelstates{t+1}\tr\rcrnthidwts\\
			&= \jacobian{H}{\hidbiases}\rcrnthidwts.\\
		\implies \rowgradient{L_{t+1}}{\modelpostmeans{t}} 
			&= \xpctt{q(\modelstates{t+1},\stateobsvs{t+1}|\modelpostmeans{t})}
					{\jacobian{H}{\hidbiases}\bigg|\modelpostmeans{t}}
				-\xpctt{q(\modelstates{t+1}|\stateobsvs{t+1},\modelpostmeans{t})}
					{\jacobian{H}{\hidbiases}\bigg|\stateobsvs{t+1},\modelpostmeans{t}}\rcrnthidwts\\
			&= \rowgradient{L_{t+1}}{\hidbiases}\rcrnthidwts.
	\end{split}
\end{equation}

Now we consider the derivatives of the hidden/output units.
From \eqn{RTRBMrecurrency}, we see that 
\begin{equation}\label{eqn:derhidsderhids}
	\jacobian{\modelpostmeans{t+1}}{\modelpostmeans{t}} = \transitionjacobian{t+1}\rcrnthidwts,
\end{equation}
where $\transitionjacobian{t+1}$ is the Jacobian of $f$, evaluated at 
$\rcrnthidwts\modelpostmeans{t} + \vishidwts\stateobsvs{t+1} + \hidbiases$.
The derivatives with respect to the parameters, for their part, are most felicitously expressed in connection with the row vector $\dfrntl{L}/\dfrntl{\modelpostmeans{t}\tr}$ (cf.\ \eqn{lossderivativeA}):
\begin{equation}\label{eqn:derhidsderparams}
	\ttlderiv{L}{\modelpostmeans{t}}\colgradient{\modelpostmeans{t}}{\rcrnthidwts}
		= \transitionjacobian{t}\tr\colttlderiv{L}{\modelpostmeans{t}}\modelpostmeans{t-1}\tr,\qquad
	\ttlderiv{L}{\modelpostmeans{t}}\colgradient{\modelpostmeans{t}}{\vishidwts}
		= \transitionjacobian{t}\tr\colttlderiv{L}{\modelpostmeans{t}}\stateobsvs{t}\tr,\qquad
	\ttlderiv{L}{\modelpostmeans{t}}\rowgradient{\modelpostmeans{t}}{\hidbiases}
		= \rowttlderiv{L}{\modelpostmeans{t}}\transitionjacobian{t}.
\end{equation}

Filling \eqnsss{derlossderparams}{derlossderhids}{derhidsderhids}{derhidsderparams} into the equations for backprop-through-time, \eqns{lossderivativeA}{backpropthroughtime}, yields
\begin{equation}\label{eqn:backpropthroughtimeB}
	\begin{split}
		\ttlderiv{L}{\modelpostmeans{t}} 
			&= \rowgradient{L_{t+1}}{\hidbiases}\rcrnthidwts
				+ \ttlderiv{L}{\modelpostmeans{t+1}} \transitionjacobian{t+1} \rcrnthidwts;\\
		\colttlderiv{L}{\rcrnthidwts}  
			&= \smplavgg{p(\stateobsvsalltime)}{
				\sum_{t=0}^{T} \colgradient{L_t}{\rcrnthidwts} 
				+ \sum_{t=0}^T \transitionjacobian{t}\tr\colttlderiv{L}{\modelpostmeans{t}}\modelpostmeans{t-1}\tr},\\
		\colttlderiv{L}{\vishidwts}  
			&= \smplavgg{p(\stateobsvsalltime)}{\sum_{t=0}^{T} \colgradient{L_t}{\vishidwts} 
				+ \sum_{t=0}^T \transitionjacobian{t}\tr\colttlderiv{L}{\modelpostmeans{t}}\stateobsvs{t}\tr},\\
		\rowttlderiv{L}{\hidbiases}  
			&= \smplavgg{p(\stateobsvsalltime)}{\sum_{t=0}^{T} \rowgradient{L_t}{\hidbiases} 
				+ \sum_{t=0}^T \rowttlderiv{L}{\modelpostmeans{t}}\transitionjacobian{t}}.\\
	\end{split}
\end{equation}
These expressions can be economized if we work with
\begin{equation}\nonumber
	 \BPTTvec{t}\tr := -\ttlderiv{L}{\modelpostmeans{t}} \transitionjacobian{t},
\end{equation}
in which case, applying also \eqn{RTRBMloss}, the complete RTRBM update equations are
\begin{equation}\label{eqn:RTRBMupdates}
	\begin{split}
		\BPTTvec{t}\tr
			&= \bigg(\rowgradient{\log q(\stateobsvs{t+1}|\modelpostmeans{t};\params)}{\hidbiases}
				+ \BPTTvec{t+1}\tr\bigg)\rcrnthidwts\transitionjacobian{t};\\
		\Delta\rcrnthidwts 
			\propto
		-\colttlderiv{L}{\rcrnthidwts} 
			&= \smplavgg{p(\stateobsvsalltime)}{\sum_{t=0}^T \bigg(
				\colgradient{\log q(\stateobsvs{t+1}|\modelpostmeans{t};\params)}{\rcrnthidwts}
				+ \BPTTvec{t}\modelpostmeans{t-1}\tr\bigg)},\\
		\Delta\vishidwts 
			\propto
		-\colttlderiv{L}{\vishidwts} 
			&= \smplavgg{p(\stateobsvsalltime)}{\sum_{t=0}^T \bigg(
				\colgradient{\log q(\stateobsvs{t+1}|\modelpostmeans{t};\params)}{\vishidwts}
				+ \BPTTvec{t}\stateobsvs{t}\tr\bigg)},\\
		\Delta\hidbiases 
			\propto
		-\rowttlderiv{L}{\hidbiases} 
			&= \smplavgg{p(\stateobsvsalltime)}{\sum_{t=0}^T \bigg(
				\rowgradient{\log q(\stateobsvs{t+1}|\modelpostmeans{t};\params)}{\hidbiases}
				+ \BPTTvec{t}\tr\bigg)},\\
		\Delta\visbiases
			\propto
		-\rowttlderiv{L}{\visbiases} 
			&= \smplavgg{p(\stateobsvsalltime)}{\sum_{t=0}^T
				\rowgradient{\log q(\stateobsvs{t+1}|\modelpostmeans{t};\params)}{\visbiases}}.
	\end{split}
\end{equation}
We have here also included the equation for updating the \emph{visible} biases, which does not depend on the backwards recursion.
In our experiments, each of the partial derivatives was approximated with contrastive divergence.

\section{Details of the experiments}\label{sec:experimentdetails}
We provide here details about the generative model used to produce the ``data distribution'' on which the temporal EFH's were trained and tested in \sctn{LDSwithPPCs}, and the testing procedure.

\subsection*{Details of the true generative model}
\newcommand{\gain}{\ensuremath{g}}
\newcommand{\dt}{\delta t}
The unobserved, linear dynamical system evolved according to
\begin{equation}\label{eqn:stateEqn}
	p(\worldstates{t+1}|\worldstates{t})
		= \nrml{A\worldstates{t}}{\cvrntransstates},
	\hspace{0.5in}
	p(\worldstates{0}) = \nrml{\vect{0}}{\cvrninitstates}.
\end{equation}
To prevent these dynamics from being well approximated by a first-order dynamical model, we made two choices:
(1) The state transition matrix was chosen to correspond to the discrete-time approximation to a(n undriven) damped harmonic oscillator, i.e., $m\ddot\worldstatevar + c\dot\worldstatevar + k\worldstatevar = 0$,
\begin{equation}\nonumber
	A = \begin{bmatrix}1 & \dt \\ -\frac{k}{m}\dt & 1-\frac{c}{m}\dt  \end{bmatrix},
\end{equation}
with mass $m = 5$ kg, viscous damping $c = 0.25$ Ns/m, ideal-spring stiffness $k = 3$ N/m, and sampling interval $\dt = 0.05 s$.
This makes the system stable and oscillatory (which first-order models cannot be).
(2) $\cvrntransstates$ was chosen so that the Hankel singular values \citep[see][]{Scherpen2011} for the system, with output matrix 
$C = \begin{bmatrix} 1 & 0\end{bmatrix}$ and input matrix set by $\cvrntransstates$, were within one order of magnitude of each other; that is, ensuring that the transfer function from noise to position had roughly equal power in all modes.
This was achieved with $\cvrntransstates = \text{diag}(\begin{bmatrix} 5\text{\sc{e}-}7, & 5\text{\sc{e}-}5\end{bmatrix})$.

The current (time $t$) position was noisily encoded in the spike counts of a population of fifteen neurons, whose Gaussian-shaped tuning curves ($\tc_i$) evenly tile the space.
Spike counts were drawn from (conditionally) independent Poisson distributions:
\begin{equation}\label{eqn:populationDataGeneration}
	p(\stateobsvs{t}|\worldstates{t},\gain_t) 
		= \prod_{i=1}^{15} \Pois{\stateobsv{i,t}|\gain_t\tc_i(\worldstate{t})}.
\end{equation}
Here $\gain_t$ is the ``population gain,'' a scaling factor for the mean spike counts \citep{Ma2006}.
Because the signal-to-noise ratio increases with mean for Poisson random variables, it essentially scales (linearly) the reliability of the population.
Thus, in order to model instant-to-instant changes in sensory reliability, the gain was chosen independently and uniformly:
\begin{equation}\label{eqn:gainDstrbs}
	p(\gain_t) = \Uniform{6.4}{9.6}.
\end{equation}
The gains are unobserved.
The joint distribution of the states and their observations and the gains is the product of \eqn{stateEqn}--\eqn{gainDstrbs}, multiplied across all time.
This distribution is consistent with the generic dynamical generative model in the first two rows of \subfig{HMMaugmented}.

In accordance with the broad tuning of higher sensory areas, the ``standard deviation,'' $\tcstd$, of the tuning curves,
\begin{equation}\nonumber
	\tc_i(x) = \exp\bigg\{ -\frac{(x-\tcmu_i)^2}{2\tcvar} \bigg\},
\end{equation}
was chosen so that the full-width at half maximum is one-sixth of the total space, for all preferred stimuli $\tcmu_i$.

This system is input-output stable (eigenvalues of the state-transition matrix are within the unit circle), but trajectories are nevertheless unbounded, since the input noise is unbounded (normally distributed).
Thus, the (rare) stimuli that leave the space spanned by the tuning curves were simply ``wrapped'' onto the opposite side.

Although we have written the initial-state distribution as Gaussian, the initial position was in fact drawn from a uniform distribution across the space spanned by the Poisson ``neurons.''
For EM learning, this was treated as an infinite-covariance Gaussian centered in the middle of joint space.
The initial velocity was normally distributed very tightly about zero, so all movement of the oscillator was essentially due to initial displacement.

\subsection*{Details of the testing procedure}
Models were tested on 40000 vectors, consisting of 40 trajectories of 1000 time steps apiece.
For each trajectory, the vector of hidden means was computed from the current (time $t$) inputs according to \eqn{rcrntdstrb}.
Then the position, $\Worldstate{t}$, was decoded from the hidden-unit means, $\modelpostmeans{t}$, using a scheme we have proposed previously for similar but static models \citep{Makin2013b}.
First, these activities were passed back down through the harmonium (i.e., through the weight matrix and element-wise nonlinearity) into the 15 ``sensory'' (i.e., the Poisson) units; and then the center of mass of these units was computed.
Since the center of mass is the maximum-likelihood estimator for such a population \citep{Makin2013b}, this is sensible, although not provably optimal, so the model performances presented in \fig{ErrorStatsVsNumHiddens} is in some sense a lower bound.
However, since the same decoding technique was used on all models, the comparisons between them are valid.

The same data were used for all tests in \fig{ErrorStatsVsNumHiddens}, so all variation across models and benchmarks results from the training alone.

\vskip 0.2in
\bibliography{\bibsdir/rEFHarXiv}

\end{document}